%% file: main.tex
\documentclass[10pt,twocolumn,letterpaper]{article}

\usepackage[pagenumbers]{cvpr}              %

\input{preamble}

\definecolor{cvprblue}{rgb}{0.21,0.49,0.74}
\usepackage[pagebackref,breaklinks,colorlinks,citecolor=cvprblue]{hyperref}

\def\paperID{91} %
\def\confName{3DV\xspace}
\def\confYear{2024\xspace}

\title{S4C: \textbf{S}elf-\textbf{S}upervised \textbf{S}emantic \textbf{S}cene \textbf{C}ompletion with Neural Fields}

\author{Adrian Hayler$^{\text{*},1}$ \hspace{0.3cm} Felix Wimbauer$^{\text{*},1,2}$ \hspace{0.3cm} Dominik Muhle$^{1,2}$ \\ Christian Rupprecht$^3$ \hspace{0.3cm} Daniel Cremers$^{1,2,3}$\\
$^1$Technical University of Munich \hspace{1cm} $^2$MCML \hspace{1cm} $^3$University of Oxford\\
{\tt\small \{a.hayler, felix.wimbauer, dominik.muhle, cremers\}@tum.de \hspace{.5cm} chrisr@robots.ox.ac.uk}
}

\newcommand\blfootnote[1]{%
  \begingroup
  \renewcommand\thefootnote{}\footnote{#1}%
  \addtocounter{footnote}{-1}%
  \endgroup
}

\begin{document}

\maketitle

\input{figures/teaser}
\input{sec/0_abstract}
\blfootnote{$^{\text{*}}$ equal contribution. \href{https://ahayler.github.io/publications/s4c/}{Project page}}

\input{sec/1_intro}

\input{sec/2_related_work}

\input{sec/3_method}

\input{sec/4_experiments}

\input{sec/5_discussion_and_conclusion}

\footnotesize{
\paragraph{Acknowledgements.}
This work was supported by the ERC Advanced Grant SIMULACRON, by the Munich Center for Machine Learning and by the EPSRC Programme Grant VisualAI EP/T028572/1.
C.\ R.\ is supported by VisualAI EP/T028572/1 and ERC-UNION-CoG-101001212.}

\newpage
{
    \small
    \bibliographystyle{ieeenat_fullname}
    \bibliography{main}
}

\appendix
\input{sec/X_suppl}

\end{document}

%% file: preamble.tex
\usepackage[dvipsnames]{xcolor}

\usepackage[T1]{fontenc}
\usepackage{acronym}
\usepackage{caption}
\usepackage{subcaption}
\usepackage{cuted}
\usepackage{adjustbox}
\usepackage{pifont}
\usepackage{algpseudocode}
\usepackage{algorithm}

\newcommand{\cmark}{\ding{51}}%
\newcommand{\xmark}{\ding{55}}%

\acrodef{ssc}[SSC]{semantic scene completion}
\acrodef{sc}[SC]{scene completion}
\acrodef{bts}[BTS]{BehindTheScenes}
\acrodef{mlp}[MLP]{multi-layer perceptron}

\usepackage{balance}

\definecolor{roadcolor}{RGB}{234,51,246}
\definecolor{sidewalkcolor}{RGB}{68,8,72}
\definecolor{parkingcolor}{RGB}{241,156,249}
\definecolor{othergroundcolor}{RGB}{160,32,76}
\definecolor{buildingcolor}{RGB}{246,202,69}
\definecolor{carcolor}{RGB}{111,149,238}
\definecolor{truckcolor}{RGB}{74,32,172}
\definecolor{bicyclecolor}{RGB}{136,227,242}
\definecolor{motorcyclecolor}{RGB}{37,59,146}
\definecolor{othervehiclecolor}{RGB}{96,81,242}
\definecolor{vegetationcolor}{RGB}{79, 173, 50}
\definecolor{trunkcolor}{RGB}{126, 65, 22}
\definecolor{terraincolor}{RGB}{171, 238, 105}
\definecolor{personcolor}{RGB}{234, 60, 49}
\definecolor{bicyclistcolor}{RGB}{234, 66, 195}
\definecolor{motorcyclistcolor}{RGB}{138, 42, 90}
\definecolor{fencecolor}{RGB}{238, 128, 69}
\definecolor{polecolor}{RGB}{252, 241, 161}
\definecolor{trafficsigncolor}{RGB}{233, 51, 35}
\definecolor{other-struct.color}{RGB}{255, 150, 0}
\definecolor{other-objectcolor}{RGB}{50, 255, 255}
\definecolor{lane-markingcolor}{RGB}{150, 255, 170}
\definecolor{color1}{RGB}{176, 36, 24}
\definecolor{color2}{RGB}{119,185,0}
\definecolor{color3}{RGB}{0, 0, 200}
\definecolor{colorofteaser}{RGB}{176, 36, 24}

\usepackage{colortbl}
\definecolor{mygray}{gray}{0.9}

%% file: figures/teaser.tex
\begin{strip}
\vspace{-1.6cm}
\centering
\captionsetup{type=figure}
\includegraphics[trim={0cm 0cm 0cm 0cm},clip,width=\linewidth]{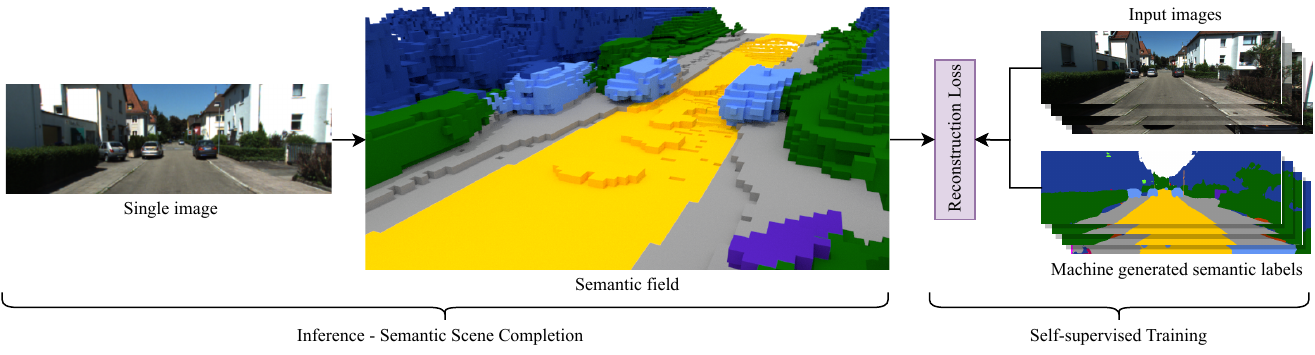}
\captionof{figure}{
\textbf{S4C, predicting a semantic field from a single image.} S4C is the first fully self-supervised \acf{ssc} method that was trained on image data alone (and camera poses). Our method predicts a volumetric scene representation from a single image, capturing geometric and semantic information even in occluded regions. In contrast to previous \ac{ssc} methods, we do not require any 3D ground truth information, allowing the use of image-only datasets for training. Despite the lack of ground truth data, our method is competitive to supervised methods with only small differences in performance. 
For further results, please check out the \textbf{video in the supplementary material}.
}
\label{fig:teaser}
\vspace{-0.2cm}
\end{strip}

%% file: sec/0_abstract.tex
\begin{abstract}
\vspace{-0.2cm}
3D semantic scene understanding is a fundamental challenge in computer vision.
It enables mobile agents to autonomously plan and navigate arbitrary environments. 
\Ac{ssc} formalizes this challenge as jointly estimating dense geometry and semantic information from sparse observations of a scene.
Current methods for \ac{ssc} are generally trained on 3D ground truth based on aggregated LiDAR scans.
This process relies on special sensors and annotation by hand which are costly and do not scale well.
To overcome this issue, our work presents the first self-supervised approach to \Ac{ssc} called \textbf{S4C} that does not rely on 3D ground truth data.
Our proposed method can reconstruct a scene from a single image and only relies on videos and pseudo segmentation ground truth generated from off-the-shelf image segmentation network during training.
Unlike existing methods, which use discrete voxel grids, we represent scenes as implicit \textit{semantic fields}.
This formulation allows querying any point within the camera frustum for occupancy and semantic class.
Our architecture is trained through rendering-based self-supervised losses.
Nonetheless, our method achieves performance close to fully supervised state-of-the-art methods.
Additionally, our method demonstrates strong generalization capabilities and can synthesize accurate segmentation maps for far away viewpoints.

\end{abstract}

\vspace{-0.75cm}

%% file: sec/1_intro.tex
\section{Introduction}
\label{sec:intro}

A plethora of tasks require holistic 3D scene understanding. Obtaining an accurate and complete representation of the scene, both with regard to geometry and high-level semantic information, enables planning, navigation, and interaction. Obtaining this information is a field of active computer vision research that has become popular with the \acf{ssc} task \cite{song2017semantic}. \ac{ssc} jointy infers the scene geometry and semantics in 3D space from limited observations \cite{li2023voxformer}. 

Current approaches to \ac{ssc} either operate on Lidar scans \cite{song2017semantic, roldao2020lmscnet} or image data \cite{cao2022monoscene, li2023voxformer, sima2023_occnet, miao2023occdepth, li2023fb} as input. 
Generally, these methods predict discrete voxel grids that contain occupancy and semantic class information.
They are trained on 3D ground truth aggregated from numerous annotated Lidar scans.
LiDAR-based methods perform better on the task of \ac{ssc} but depend on costly sensors compared to cameras \cite{li2023sscbench}. 
Cameras are readily available and offer a dense representation of the world. 
However, bridging the gap between 2D camera recordings and 3D voxel grids is not straight-forward. 
MonoScene \cite{cao2022monoscene} uses line-of-sight projection to lift 2D features into 3D space. 
However, this disregards information for occluded and empty scene regions \cite{li2023voxformer}. %
VoxFormer \cite{li2023voxformer} uses a transformer network to simultaneously predict geometry and semantic labels starting from a few query proposals on a voxel grid.

Recently, neural fields 
have emerged as a versatile representation for 3D scenes. 
Here, an MLP learns a mapping from encoded coordinates to some output.
While initially focused on geometry and appearance, they have since progressed to also incorporate semantic information \cite{Zhi2021semanticNERF, siddiqui2023panoptic, fu2022panoptic, KunduCVPR2022PNF, wang2022dm}. 
One of the major drawbacks of neural fields is that they rely on test time optimization. 
The network weights are trained by reconstructing different input views of the scene via volume rendering.
To enable generalization on scene geometry and appearance, some methods \cite{yu2021pixelnerf, wimbauer2023behind} proposed to condition neural fields on pixel-aligned feature maps predicted by trainable image encoders.

In this work, we introduce the first self-supervised approach to semantic scene completion (\acs{ssc}).
Instead of predicting a voxel grid, our method infers a 3D semantic field from a single image.
This field holds density and semantic information and allows for volume rendering of segmentation maps and color images (via image-based rendering).
By applying reconstruction losses on the rendered output in 2D, we learn the geometry and semantics of the 3D scene.
We train our approach on multiple views from the videos captured by a multi-cam setup mounted on a moving vehicle.
To make our method as general as possible, we rely on segmentation maps generated by an off-the-shelf image segmentation network rather than hand-annotated ground truth.
In order to learn geometry and semantics in the entire camera frustum, we sample frames from the videos at random time offsets to the input image.
As the vehicle moves forward, the different cameras capture many areas of the scene, especially those that are occluded in the input image.
Our formulation does not require any form of 3D supervision besides camera poses.

We use the KITTI-360 \cite{liao2022kitti} dataset for training and measure the performance of our proposed approach on the new SSCBench \cite{li2023sscbench} dataset, which is defined on top of KITTI-360.
Both qualitative and quantitative results show that our method achieves comparable results to fully-supervised approaches for \acs{ssc}.
Further, we demonstrate the beneficial effects of our different loss components and the random frame sampling strategy.
Finally, we also test the unique ability of our method to synthesize high-qualtiy segmentation maps from novel views.

Our \textbf{contributions} can be summarized as follows:
\begin{itemize}
    \item We propose the \textbf{first \ac{ssc} training using self-supervision from images without the need for 3D ground truth data}. %
    \item We achieve close-to state-of-the-art performance compared to fully supervised \acs{ssc} methods.
    \item Our method can synthesize high-quality segmentation maps from novel views.
    \item We release our code upon acceptance to further facilitate research into \ac{ssc}.
\end{itemize}

%% file: sec/2_related_work.tex
\section{Related Work}
\label{sec:related_work}
In the following, we will introduce relevant literature to our proposed method. For \acf{ssc}, we will focus the discussion of related work on camera-based approaches and introduce LiDAR-based methods only briefly. We refer the interested reader to \cite{roldao20223d} for a broader overview of the topic of \ac{ssc}. 

\subsection{Single Image Scene Reconstruction}
Scene reconstruction refers to the task of estimating 3D geometry from images. While it has been a topic of active research for two decades \cite{hartley2003multiple}, the introduction of NeRFs \cite{mildenhall2020nerf} has led to renewed interest in this area. An overview can be found in \cite{han2021sota_overview}. In the following, we restrict our review to single-view methods and their distinction from monocular depth estimation methods. Monocular depth estimation \cite{godard2017unsupervised, godard2019digging, spencer2020defeat, lyu2021hr, zhou2021diffnet} reconstructs a 3D environment by predicting a per-pixel depth value. Ground-truth supervision \cite{eigen2014depth, liu2015learning, aich2021bidirectional, fu2018deep, lee2021patch, li2022depthformer, li2022binsformer, wimbauer2021monorec}, reconstruction losses \cite{zhou2017unsupervised, zhan2018unsupervised, godard2017unsupervised, godard2019digging}, or a combination thereof \cite{kuznietsov2017semi, yang2018deep} have been used to train these networks. In contrast to depth estimation and its restriction to visible surfaces, scene reconstruction aims to also predict geometry in occluded regions. PixelNeRF \cite{yu2021pixelnerf}, a NeRF variant with the ability to generalize to different scenes, can predict free space in occluded scene regions only from a single image. As an extension, SceneRF \cite{cao2023scenerf} uses a probabilistic sampling and a photometric reprojection loss to additionally supervise depth prediction to improve generalization. BTS \cite{wimbauer2023behind} combines ideas from generalizable NeRF and self-supervised depth estimation and achieves very accurate geometry estimation, even for occluded regions. 
In contrast to our approach, the above-mentioned methods do not consider semantics and are therefore not suited for \acs{ssc}.
Another line of work for scene reconstruction leverages massive data to learn object shape priors \cite{choy20163d, tulsiani2017multi, fan2017point, yan2016perspective}.

\subsection{3D Semantic Segmentation}

Given a 3D model, such as mesh, semantic multi-view fusion models \cite{hermans2014dense, kundu2020virtual, armeni20193d, mccormac2017semanticfusion, mascaro2021diffuser, zhang2019large} project segmentation masks from images into the 3D geometry. Implicit representations such as NeRFs have become popular for semantic 3D modeling \cite{Zhi2021semanticNERF, fu2022panoptic, vora2021nesf, KunduCVPR2022PNF, wang2022dm} to ensure multi-view consistency of segmentation masks. \cite{ siddiqui2023panoptic} extends this idea to the panoptic segmentation task. The works of ConceptFusion \cite{jatavallabhula2023conceptfusion} and OpenScene \cite{peng2023openscene} propose open vocabulary scene understanding by fusing open vocabulary representations into 3D scene representations allowing for segmentation as a downstream task. 

In contrast to image segmentation, Lidar segmentation works with 3D data to assign semantic labels to point cloud data. Unlike images, point clouds are a sparse and unordered data collection. To address this different data modality, Lidar segmentation either use point-based \cite{qi2017pointnet, qi2017pointnet++, wang2019dynamic, he2019geonet, zamorski2020adversarial}, voxel-based \cite{maturana2015voxnet, qi2016volumetric, wang2019normalnet}, or projection-based methods \cite{wu2018squeezeseg, beltran2018birdnet, milioto2019rangenet++, alnaggar2021multi}.

\subsection{Semantic Scene Completion}
\Acf{ssc} extends the task of completing the scene geometry by jointly predicting scene semantics.
It was first introduced in \cite{song2017semantic} and has gained significant attention in recent years \cite{roldao20223d}. 
This contrasts the separate treatment of the tasks in early works \cite{gupta2013perceptual, thrun2005shape}. 
The first approaches on \ac{ssc} either focused on indoor settings from image data \cite{zhang2018efficient,  zhang2019cascaded, liu2018see, li2019depth, li2019rgbd, li2020anisotropic, chen20203d, cai2021semantic} or outdoor scenes with LiDAR-based methods \cite{roldao2020lmscnet, cheng2021s3cnet, yan2021sparse, li2021semisupervised, rist2021semantic}. 
MonoScene \cite{cao2022monoscene} was the first to present a unified camera approach to indoor and outdoor scenarios using a line-of-sight projection and a novel frustum proportion loss. VoxFormer \cite{li2023voxformer} uses deformable cross-attention and self-attention on voxels from image features. OccDepth \cite{miao2023occdepth} utilizes stereo images and stereo depth supervision. 
Another line of work uses birds-eye-view and temporal information for 3D occupancy prediction \cite{sima2023_occnet, li2023fb}. This idea was extended to a Tri-Perspective View in \cite{huang2023tri}. SSCNet first tackled the problem of \ac{ssc} from an image and a depth map and used a 3D convolutional network to output occupancy and labels in a voxel grid \cite{song2017semantic}. LSMCNet combines 2D convolutions with multiple 3D segmentation heads at multiple resolutions to reduce network parameters \cite{roldao2020lmscnet}. 
Overall, Lidar methods tend to outperform camera approaches on outdoor scenes.
All the above-mentioned methods require annotated 3D ground truth for training, which is costly to collect.
The need for accurate 3D ground truth data restricts the evaluation of \ac{ssc} methods to only a few datasets, such as SemanticKITTI \cite{behley2019semantickitti}. SSCBench \cite{li2023sscbench} is a recently introduced benchmark that includes annotated ground truth for \acs{ssc} on KITTI-360 \cite{liao2022kitti}, nuScenes \cite{caesar2020nuscenes}, and Waymo \cite{sun2020scalability}. 

In contrast, we present \textbf{S4C}, a fully self-supervised training strategy that allows our model to be trained from posed images only, lifting the restriction on the expensive datasets with Lidar data.

%% file: sec/3_method.tex
\section{Method}
\label{sec:method}

In the following, we describe our approach to predict the geometry and semantics of a scene from a single image $\textbf{I}_\text{I}$ to tackle the task of Semantic Scene Completion, as shown in \cref{fig:overview}. 
We first cover how we represent a scene as a continuous semantic field, and then propose a fully self-supervised training scheme that learns 3D geometry and semantics from 2D supervision only.

\input{figures/overview}

\subsection{Notation}
Let $\textbf{I}_\text{I} \in [0, 1]^{3\times H\times W} = (\mathbb{R}^3)^\Omega$ be the input image which is defined on a lattice $\Omega = \{1, \ldots, H\}\times\{1, \ldots, W\}$.
During training, we have access to $N = \{\textbf{I}_1, ..., \textbf{I}_n\}$ additional views of the scene beside the input image.
Through an off-the-shelf semantic segmentation network $\Phi(\textbf{I})$, we obtain segmentation maps $\textbf{L}_i \in \{0, \dots , c-1\}^{\Omega}$ for all images $\textbf{I} \in \{\textbf{I}_\textbf{I}\} \cup N$. 
$c$ denotes the number of different classes. 
Camera poses and projection matrix of the images are given as $T_i \in \mathbb{R}^{4\times 4}$ and $K_i \in \mathbb{R}^{3 \times 4}$, respectively. 
Points in world coordinates are denoted as $\textbf{x} \in \mathbb{R}^3$. Projection into the image plane of camera $k$ is given by $\pi_k(\textbf{x}) = K_k T_k \textbf{x}$ in homogeneous coordinates. 

\subsection{Predicting a Semantic Field}

Current approaches to \ac{ssc} typically involve the prediction and manipulation of discrete voxel grids, which come with various limitations. Firstly, these grids have restricted resolution due to their cubic memory requirements and processing constraints.
Second, when reconstructing a scene from an image, voxels are not aligned with the pixel space. 
Consequently, methods have to rely on complex multi-stage approaches to lift information from 2D to 3D \cite{cao2022monoscene, li2023voxformer}.

As an alternative, we propose a simple architecture that predicts an \textit{implicit} and \textit{continuous} \textbf{semantic field} to overcome these shortcomings. \cite{Zhi2021semanticNERF}
A semantic field maps every point $\textbf{x}$ within the camera frustum to both volumetric density $\sigma \in [0, 1]$ and a semantic class prediction $l \in \{0, \dots , c-1\}$.
Inspired by \cite{yu2021pixelnerf, wimbauer2023behind}, we use a high capacity encoder-decoder network to predict a dense pixel-aligned feature map $\textbf{F} \in (\mathbb{R}^3)^\Omega$ from the input image $\textbf{I}_\textbf{I}$.
The feature $f_\textbf{u}$ at pixel location $\textbf{u}$ describes the semantic and geometric structure of the scene captured along a ray through that pixel.
We follow \cite{wimbauer2023behind} and do not store color in the neural field. 
This improves generalization capabilities and robustness.

To query the semantic field at a specific 3D point $\textbf{x} \in \mathbb{R}^3$ within the camera frustum, we first project $\textbf{x}$ onto the camera plane to obtain the corresponding pixel location $\textbf{u}$.
First, $\textbf{x}$ is projected onto the image plane $\textbf{u}_\text{I} = \pi_{\text{I}}(\textbf{x})$. The corresponding feature vector is extracted from the feature map with bilinear interpolation $f_{\textbf{u}} = \textbf{F}(\textbf{u})$. Together with positional encodings $\gamma(d)$ for distance $d$ \cite{mildenhall2020nerf} and pixel position $\gamma(\textbf{u}_\text{I})$, the feature vectors is then decoded to the density
\begin{equation}
    \sigma_\textbf{x} = \phi_{\text{D}}(f_{\textbf{u}_\text{I}}, \gamma(d), \gamma(\textbf{u}_\text{I}))
\end{equation}
and semantic prediction
\begin{equation}
    l_{\textbf{x}} = \phi_{\text{S}}(f_{\textbf{u}_\text{I}}, \gamma(d), \gamma(\textbf{u}_\text{I})) \,.
\end{equation}
Both $\phi_{\text{D}}$ and $\phi_{\text{S}}$ are small \ac{mlp} networks.
Note that $\phi_{\text{S}}$ predicts semantic logits. 
To obtain a class distribution or label, we apply $\underset{c}{\operatorname{Softmax}}\left(l_{\textbf{x}}\right)$ or $\underset{c}{\arg \max} \left(l_{\textbf{x}}\right)$, respectively.

\subsection{Volumetric Semantic Rendering}
The goal of this paper is to develop a method to perform 3D \acs{ssc} from a single image while being trained from 2D supervision alone. 
The continuous nature of our scene representation allows us to use volume rendering \cite{kajiya1984ray, max1995optical} to synthesize high-quality novel views. 
As shown in \cite{Zhi2021semanticNERF}, volumetric rendering can be extended from color to semantic class labels. 
The differentiable rendering process allows us to back-propagate a training signal from both color and semantic supervision on rendered views to our scene representation.

To render segmentation masks from novel viewpoints, we cast rays from the camera for every pixel. Along each ray, we integrate the semantic class labels over the probability of the ray ending at a certain
distance. 
To approximate this integral, density $\sigma_{\textbf{x}_i}$ and label $l_{\textbf{x}_i}$ are
evaluated at $m$ discrete steps $\textbf{x}_i$ along the ray.

Since we consider segmentation in 3D space, we apply $\operatorname{Softmax}$ normalization at every query point individually, \textit{before} we integrate along the ray.
The intuition behind this is that regions with low density should not be able to ``overpower'' high-density regions by predicting very high scores for classes.
Thus, this technique makes rendering semantics from the neural field more robust. \cite{siddiqui2023panoptic}

\begin{equation}
    \alpha_i = 1 - \exp(- \sigma_{\textbf{x}_i} \delta_i)\quad\quad
    T_i = \prod_{j=1}^{i-1} (1 - \alpha_j)%
\end{equation}
\begin{equation}
    \hat{l} = \sum_{i=1}^{m}T_i \alpha_i \cdot \underset{c}{\operatorname{Softmax}}\left(l_{\textbf{x}_i}\right)
\end{equation}
Here, $\delta_i$ denotes the distance between the points $\textbf{x}_i$ and $\textbf{x}_{i + 1}$, $\alpha_i$ the probability of the ray ending between the points $\textbf{x}_i$ and $\textbf{x}_{i + 1}$, and $T_i$ the probability of $\textbf{x}_i$ being not occluded and therefore visible in the image. $\hat{l}$ is the final, normalized distribution predicted for a pixel.
We construct a per-pixel depth $d_i$, the distance between $\textbf{x}_i$, and the ray origin using the expected ray termination depth.
\begin{equation}
    \hat{d} = \sum_{i=1}^{m}T_i \alpha_i d_i
\end{equation}

From the semantic field, we can also synthesize novel appearance views by applying image-based rendering \cite{wimbauer2023behind}. 
Image-based rendering does not predict color information from a neural field but rather queries the color from other images. 
We sample the color by projecting point $\textbf{x}_i$  into frame $k$ to the pixel position $\textbf{u}_{i,k} = \pi_k(\textbf{x}_i)$ and then bi-linearly interpolating the color value $c_{i, k} = \textbf{I}_k(\textbf{u}_{i,k})$ at the projected pixel position. 
The color sample can come from any other image $k$ except the one we want to render. 
With volumetric rendering, we obtain the pixel color with queries from image $k$ as
\begin{equation}
    \hat{c}_k = \sum_{i=1}^{m}T_i \alpha_i c_{i, k} \,,
\end{equation}
where we obtain a different color prediction from every frame $k$.

\subsection{Training for Semantic Multi-View Consistency}

All existing methods for \acs{ssc} rely on 3D ground truth data for training.
Such data is generally obtained from annotated LiDAR scans, which are very difficult and costly to acquire.
In contrast to 3D data, images with semantic labels are abundantly available. 
We propose to leverage this available 2D data to train a neural network for 3D semantic scene completion.
To make our approach as general as possible, we use pseudo-semantic labels generated from a pre-trained semantic segmentation network. 
This allows us to train our architecture only from posed images without the need for either 2D or 3D ground truth data in a fully self-supervised manner. 

In this paper, we consider \acs{ssc} in an autonomous driving setting. 
Generally, there are forward and sideways-facing cameras mounted on a car that is moving.
We train our method on multiple posed images. In addition to the source image $\textbf{I}_\textbf{I}$ from which a network produces the feature map $\textbf{F}$, frames $\textbf{I}_i$ are aggregated from the main camera, stereo, and side-view cameras over multiple timesteps of a video sequence.

We randomly sample a subset of the available frames and use them as reconstruction targets for novel view synthesis.
Our pipeline reconstructs both colors and semantic labels - based on our semantic field and color samples from other frames.
The discrepancy between the pseudo-ground truth semantic masks and the image is used as the training signal.

As supervision from a view only gives training signals in areas that are observed by this camera, it is important to select training views strategically.
Especially sideways-facing camera views give important cues for regions that are occluded in the input image.
In order to best cover the entire camera frustum we aim to reconstruct, we, therefore, select sideways-facing views with a random offset to the input image for training.
This increases the diversity and improves prediction quality especially for further away regions, which are often difficult to learn with image-based reconstruction methods.
 
In practice, we only reconstruct randomly sampled patches $P_i$ that we reconstruct from different render frames $k$ as $\hat{P}_{i, k}$ for color. For semantic supervision, we reconstruct the semantic labels $S_i$ as $\hat{S}_{i}$.
To train for \ac{ssc}, \ie scene reconstruction and semantic labelling, we use a combination of semantic and photometric reconstruction loss. 
While the photometric loss gives strong training signals for the general scene geometry, the semantic loss is important for differentiating objects and learning rough geometry.
Furthermore, it guides the model to learn sharper edges around objects. %

We use weighted binary cross-entropy loss to apply supervision on the density field from our pseudo ground truth segmentation labels $\mathcal{L}_i$.
\begin{equation}
    \mathcal{L}_\text{sem} = \text{BCE}\left(S_i, \hat{S}_i\right)
    \label{eq:loss_sem}
\end{equation}

The photometric loss employs a combination of L1, SSIM \cite{wang2004image} and an edge-aware smoothness (eas) term loss as proposed in \cite{godard2019digging}. We follow the strategy of \cite{wimbauer2023behind, godard2019digging} to take only the per-pixel \textit{minimum} into account when aggregating the costs. 
\begin{equation}
    \mathcal{L}_\text{ph} = \min_{k\in N_{\text{render}}} \left(\lambda_\text{L1} \text{L1}(P_i, \hat{P}_{i, k}) + \lambda_\text{SSIM} \text{SSIM}(P_i, \hat{P}_{i, k})\right)
\label{eq:loss_ph}
\end{equation}
\begin{equation}
    \mathcal{L}_\text{eas} = \left|\delta_x d^\star_i\right|e^{-\left|\delta_x P_i\right|} + \left|\delta_y d^\star_i\right|e^{-\left|\delta_y P_i\right|}
\label{eq:loss_eas}
\end{equation}

Our final loss is then computed as a weighted sum:
\begin{equation}
    \mathcal{L} = \mathcal{L}_\text{sem} + \lambda_\text{ph}  \mathcal{L}_\text{ph} + \lambda_\text{eas}  \mathcal{L}_\text{eas}
\label{eq:loss}
\end{equation}

%% file: figures/overview.tex
\begin{figure*}
\centering
\captionsetup{type=figure}
\includegraphics[trim={0cm 0cm 0cm 0cm},clip,width=\linewidth]{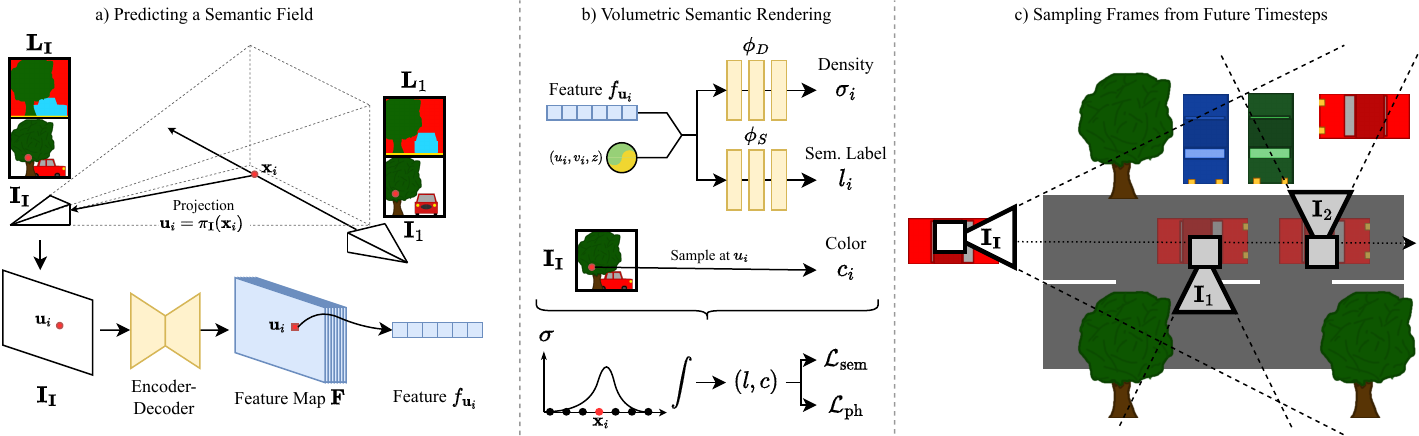}
\captionof{figure}{
\textbf{Overview.} \textbf{a)} From an input image $\textbf{I}_\textbf{I}$, an encoder-decoder network predicts a pixel-aligned feature map $\textbf{F}$ describing a semantic field in the frustum of the image. The feature $f_{\textbf{u}_i}$ of pixel $\textbf{u}_i$ encodes the semantic and occupancy distribution on the ray cast from the optical center through the pixel. \textbf{b)} The semantic field allows rendering novel views and their corresponding semantic segmentation via volumetric rendering. A 3D point $\textbf{x}_i$ is projected into the input image and therefore $\textbf{F}$ to sample $f_{\textbf{u}_i}$. Combined with positional encoding of $\textbf{x}_i$, two MLPs decode the density of the point $\sigma_i$ and semantic label $l_i$, respectively. The color $c_i$ for novel view synthesis is obtained from other images via color sampling. \textbf{c)} To achieve best results, we require training views to cover as much surface of the scene as possible. Therefore, we sample side views from random future timesteps, that observe areas of the scene that are occluded in the input frame.
}
\label{fig:overview}
\end{figure*}

%% file: sec/4_experiments.tex
\section{Experiments}
\label{sec:experiments}

To demonstrate the capabilities of our proposed method, we conduct a wide range of experiments. 
First, we evaluate our method using the new SSCBench dataset \cite{li2023sscbench} on KITTI-360 \cite{liao2022kitti} and achieve performance that closely rivals state-of-the-art, but supervised,  methods.
We also conduct ablation studies to justify our design choices and to demonstrate synergistic effects between semantic and geometric training. %
Second, we show that our method can synthesize high-quality segmentation maps for novel viewpoints.

\subsection{Implementation Details}
For the architecture, we rely on a ResNet-50 \cite{he2016deep} pre-trained on ImageNet as the backbone and a prediction head based on \cite{godard2019digging}.
The feature vectors $f_\textbf{u}$ have a dimension of $64$. 
Both MLPs $\phi_D$ and $\phi_S$ are very lightweight with two fully-connected layers of $64$ hidden nodes each.
They do not need more capacity, as all information is already contained in the feature vector $f_\textbf{u}$ and the MLPs are tasked to decode the contained information at a certain distance.

We implement our architecture entirely on PyTorch \cite{paszke2019pytorch} and train with a batch size of 16 on a single Nvidia A40 GPU with 48GB memory.
For every input image, we sample $32$ patches of size $8\times8$ for RGB and semantics reconstruction each.
For further technical details, please refer to the supplementary material.

\subsection{Data}

For training and testing our method, we use the established KITTI-360 \cite{liao2022kitti} dataset, which consists of video sequences captured by multiple cameras mounted on top of a moving vehicle.
Besides a pair of forward-facing stereo cameras, KITTI-360 provides recordings from two fisheye cameras facing sideways left and right. 
The fisheye cameras allow us to gather geometric and semantic information in parts of the scene occluded in the source view. 
We are interested in an area approximately 50 meters in front of the car. 
Based on this and considering an average speed of 30-50kph, we sample fisheye views between 1 to 4 seconds into the future. 
The recordings in KITTI-360 have a frame rate of $10\text{Hz}$ which translates to an offset of 10 - 40 timesteps.
In total, we use a total of eight images per sample during training: Four forward-facing views (out of which one is the input image), and four sideways-facing views.

To generate the pseudo segmentation labels, we run the off-the-shelf segmentation network Panoptic-Deeplab \cite{cheng2020panoptic} trained on Cityscapes \cite{cordts2016cityscapes}.

\subsection{SSC Performance on SSCBench}

To compare the performance of our method with supervised approaches, we evaluate the predicted semantic fields on the new SSCBench dataset \cite{li2023sscbench} which is defined on KITTI-360 \cite{liao2022kitti}. 
This dataset aggregates Lidar scans over entire driving sequences and builds voxel grids with occupancy and semantic information. 
These voxel grids can be used to compute both occupancy (geometry) and semantic-focused performance metrics.
SSCBench follows the setup of SemanticKITTI \cite{behley2019semantickitti} to use a voxel resolution of 0.2m on scenes of size $51.2m\times 51.2m\times6.4m$, $256\times256\times32$ voxel volumes. 
Evaluation of SSCBench happens at three scales of $12.8m\times12.8m\times6.4m$, $25.6m\times25.6m\times6.4m$, and $51.2m\times51.2m\times6.4m$.
SSCBench also provides invalid masks for voxels that do not belong to the scene. 
As these are quite coarse, we use slightly refined invalid masks.
For fairness, we rerun all related methods on this refined data and observe a minor performance increase for all approaches.
For further details, please refer to the supplementary material.

It is important to note that evaluating on SSCBench means bridging a non-trivial domain gap for our method.
Our approach is trained via 2D supervision, which means that scene geometry must be pixel-level accurate. 
Occupancy is predicted for areas \textit{within} objects.
On the other hand, SSCBench ground truth was collected by aggregating Lidar point clouds. 
Here, a voxel is considered occupied when a Lidar measurement point lies within the voxel, but Lidar measurement points only capture the \textit{surface} of an object.
Therefore, the voxel ground truth of SSCBench tends to grow objects in size. %

We use two techniques during the discretization of our implicit semantic field into a voxel grid to best align our predictions with the way the ground truth was captured.
First, we leverage the unique advantage of our architecture to be not bound to the coarse resolution of discrete voxel grids.
For every voxel, we query \textit{multiple} different points distributed evenly within the voxel.
The final prediction is obtained by taking the maximum occupancy value among the points and weighting the class predictions accordingly.
The intuition behind this is that we want to check whether there is a surface anywhere in the voxel.
However, we observe that volumetric rendering encourages the occupancy to blur at object borders.
In a second step, we, therefore, perform a neighbourhood check. 
A voxel is considered occupied when volumetric occupancy is observed in at least one of the immediate neighbouring voxels.

Our method is the first to approach \acs{ssc} in a self-supervised manner without relying on 3D ground truth. 
We compare against several fully supervised, state-of-the-art approaches, namely MonoScene~\cite{cao2022monoscene}, LMSCNet~\cite{roldao2020lmscnet}, and SSCNet~\cite{song2017semantic}.
While MonoScene takes single images as input at test time, LMSCNet and SSCNet require Lidar scans even at test time.

\input{tables/sscbench_main}
As can be seen in \cref{tab:sscbench-main}, even though we tackle a significantly more challenging task, our proposed \textbf{S4C} method achieves occupancy IoU and segmentation mean IoU results that closely rival these of MonoScene~\cite{cao2022monoscene}, which is trained with annotated 3D Lidar ground truth.
Furthermore, the results are not far off from the methods that take Lidar inputs at test time.
Even though further away regions (25.6m and 51.2m) are usually more challenging for self-supervised approaches, as fewer camera views observe them, the performance of our method stays stable even when evaluating further away distances.
We attribute this to the strategy of sampling side views at a random time offset.
While the occupancy precision of our method is lower than other methods, our occupancy recall is significantly higher. 
A possible reason for this is that our method tends to place occupied voxels in areas that are unobserved and that cannot be hallucinated in a meaningful way from the observations in the image.
Methods trained on inherently more sparse voxel ground truth tend to place unoccupied voxels in such regions. 

\input{figures/ssc_main}
When visualizing the predicted voxel grids, as shown in \cref{fig:ssc_main}, we can see that our method is able to accurately reconstruct the given scene and to assign correct class labels.
The general structure of the scene is recovered at a large scale and smaller objects like cars are clearly separated from the rest of the scene.
Even for further distances, which are more difficult for camera-based methods, our method still provides reasonable reconstructions.
As mentioned before, our method can be observed to place more occupied in unseen, ambiguous regions.
Generally, the qualitative difference between the reconstructions by our method and supervised approaches is very low. 
This suggests that self-supervised training is a viable alternative to costly fully-supervised approaches to \acs{ssc}.

\subsection{Ablation studies}
\label{ssec:ablation}

To give insights into the effect of our design choices, we conduct thorough ablation studies using the SSCBench ground truth.
\input{tables/sscbench_ablation}
\input{figures/sscbench_ablation}

First, to investigate the effect of the different 2D supervision signals, we train our architecture with the different loss terms turned off and report the results in \cref{tab:sscbench-ablation}.
The photometric loss alone can already give a clear training signal and allows the network to recover accurate scene geometry.
Despite the semantic loss being very high-level and not providing a signal for smaller geometric details, it is enough to guide the network to predict rough geometry correctly. 
As shown in \cref{fig:sscbench-ablation}, the model can synthesize depth maps that clearly depict a geometric understanding of the scene.
Best occupancy results are achieved when training relies on both losses.
We hypothesize that the object boundaries in segmentation maps, which are much clearer compared to regular images, help the model to learn sharper geometry.

Second, we investigate the impact of our view sampling strategy.
When sampling sideways-facing views only at a fixed distance ($1s$ into the future), the model is not able to learn about far-away geometry.
Therefore, the performance is weaker especially when evaluating larger scenes ($25.6m$ and $51.2m$).
This effect can also be observed qualitatively in \cref{fig:sscbench-ablation}. 

\subsection{Synthesizing Segmentation Maps}

\input{tables/synth_seg}

Besides experiments for \acs{ssc}, we also analyse the effectiveness of our training with pseudo-ground truth labels from an off-the-shelf segmentation network.
To this end, we train our model with different levels of pseudo supervision, by providing pseudo segmentation maps for the input frame (one image), all forward-facing views (four images), and all forward and sideways-facing views (eight images) respectively.

We evaluate the segmentation performance for the input image against the real ground-truth segmentation provided by KITTI-360. Furthermore, we project segmentation maps 5, 10 and 15 time steps into the future using our predicted geometry. 
This tests the model's ability to reason about segmentation in 3D.
We report results in \cref{tab:synth_seg}.

Providing segmentation masks for more frames during training improves the segmentation performance in all settings. 
Given pseudo ground truth for all frames, our model is even able to improve over the pseudo segmentation ground truth it was trained with, which achieves an accuracy of $87.7\%$ against the KITTI-360 ground truth. This trend manifests when synthesizing novel segmentation views a number of timesteps away, where the discrepancy between the best and worst model configuration rises from $1.3\%$ to over $3.4\%$.

We hypothesize that this is due to two reasons:
First, the pseudo-segmentation ground truth is imperfect and contains artefacts. By forcing the model to satisfy segmentations from different views, the model automatically learns cleaner segmentation masks with fewer artefacts.
Second, by having more training views, we improve the 3D geometry, as shown in \cref{ssec:ablation}, and are able to learn better semantic labels in occluded regions. The further away we synthesize views, the more important such 3D semantic understanding is for accurate predictions.

%% file: tables/sscbench_main.tex
\begin{table*}[t]\centering
\renewcommand\tabcolsep{4pt}
\scriptsize
\begin{tabular}{l|ccc|ccc|ccc|ccc}
\toprule
\textit{Method} & \multicolumn{3}{c}{\textbf{S4C (Ours)}} & \multicolumn{3}{c}{\textbf{MonoScene}~\cite{cao2022monoscene}} & \multicolumn{3}{c}{\textbf{LMSCNet}~\cite{roldao2020lmscnet}} & \multicolumn{3}{c}{\textbf{SSCNet}~\cite{song2017semantic}} \\
\midrule
\textbf{Supervision}   & \multicolumn{3}{c}{\cellcolor{LimeGreen}\textbf{Camera only}} & \multicolumn{3}{c}{\cellcolor{Goldenrod}\textbf{Lidar training}}  & \multicolumn{6}{c}{\cellcolor{RedOrange}\textbf{Lidar training + testing}} \\
\midrule
\textbf{Range}         & \textbf{12.8m} & \textbf{25.6m} & \textbf{51.2m} & \textbf{12.8m} & \textbf{25.6m} & \textbf{51.2m} & \textbf{12.8m} & \textbf{25.6m} & \textbf{51.2m} & \textbf{12.8m} & \textbf{25.6m} & \textbf{51.2m} \\
\midrule
\textbf{IoU}           & 54.64 & 45.57 & 39.35 & 58.61 & 48.15 & 40.66 & 66.74 & 58.48 & 47.93 & 74.93 & 66.36 & 55.81 \\
\textbf{Precision}     & 59.75 & 50.34 & 43.59 & 71.79 & 67.02 & 64.79 & 80.58 & 76.75 & 76.87 & 83.65 & 77.85 & 75.41 \\
\textbf{Recall}        & 86.47 & 82.79 & 80.16 & 76.15 & 63.11 & 52.20 & 79.54 & 71.07 & 56.00 & 87.79 & 81.80 & 68.22 \\
\midrule
\textbf{mIoU}          & 16.94 & 13.94 & 10.19 & 20.44 & 16.42 & 12.34 & 22.01 & 19.81 & 15.36 & 26.64 & 24.33 & 19.23 \\
\textbf{car}           & 22.58 & 18.64 & 11.49 & 36.05 & 29.19 & 20.87 & 39.6  & 32.48 & 20.63 & 52.72 & 45.93 & 31.89 \\
\textbf{bicycle}       & 0.00  & 0.00  & 0.00  & 2.69  & 1.07  & 0.49  & 0.00  & 0.00  & 0.00  & 0.00  & 0.00  & 0.00  \\
\textbf{motorcycle}    & 0.00  & 0.00  & 0.00  & 4.70  & 1.44  & 0.59  & 0.00  & 0.00  & 0.00  & 1.41  & 0.41  & 0.19  \\
\textbf{truck}         & 7.51  & 4.37  & 2.12  & 19.81 & 14.14 & 8.48  & 0.62  & 0.44  & 0.23  & 16.91 & 14.91 & 10.78 \\
\textbf{other-vehicle} & 0.00  & 0.01  & 0.06  & 8.81  & 5.61  & 2.78  & 0.00  & 0.00  & 0.00  & 1.45  & 1.00  & 0.60  \\
\textbf{person}        & 0.00  & 0.00  & 0.00  & 2.26  & 1.30  & 0.87  & 0.00  & 0.00  & 0.00  & 0.36  & 0.16  & 0.09  \\
\textbf{road}          & 69.38 & 61.46 & 48.23 & 82.94 & 73.32 & 58.23 & 84.60 & 81.24 & 69.06 & 87.81 & 85.42 & 73.82 \\
\textbf{sidewalk}      & 45.03 & 37.12 & 28.45 & 56.51 & 43.53 & 32.70 & 60.73 & 51.28 & 36.71 & 67.19 & 60.34 & 46.96 \\
\textbf{building}      & 26.34 & 28.48 & 21.36 & 39.17 & 38.02 & 31.79 & 48.59 & 51.55 & 41.22 & 53.93 & 54.55 & 44.67 \\
\textbf{fence}         & 9.70  & 6.37  & 3.64  & 12.36 & 6.70  & 3.83  & 1.64  & 0.62  & 0.26  & 14.39 & 10.73 & 6.42  \\
\textbf{vegetation}    & 35.78 & 28.04 & 21.43 & 38.26 & 31.51 & 25.67 & 51.17 & 46.90 & 38.70 & 56.66 & 51.77 & 43.30 \\
\textbf{terrain}       & 35.03 & 22.88 & 15.08 & 38.05 & 27.30 & 19.29 & 43.23 & 32.59 & 23.54 & 43.47 & 36.44 & 27.83 \\
\textbf{pole}          & 1.23  & 0.94  & 0.65  & 10.41 & 9.25  & 7.34  & 0.00  & 0.00  & 0.00  & 1.03  & 1.05  & 0.62  \\
\textbf{traffic-sign}  & 1.57  & 0.83  & 0.36  & 9.20  & 7.98  & 5.68  & 0.00  & 0.00  & 0.00  & 1.01  & 1.22  & 0.70  \\
\textbf{other-object}  & 0.00  & 0.00  & 0.00  & 6.62  & 5.17  & 3.44  & 0.00  & 0.00  & 0.00  & 1.20  & 0.97  & 0.58  \\
\bottomrule
\end{tabular}
\caption{\textbf{Quantitative evaluation on SSCBench-KITTI-360}. We report the performances with respect to different ranges (12.8m, 25.6m, and 51.2m). We provide both \textbf{geometric} (IoU, Precision, Recall) and \textbf{semantic} (mIoU, per class IoU) metrics. As we use refined invalid masks on SSCBench, we rerun all methods with the provided checkpoints. MonoScene~\cite{cao2022monoscene} is trained with Lidar but also uses a single image at test time. LMSCNet~\cite{roldao2020lmscnet} and SSCNet~\cite{song2017semantic} are trained on Lidar data and require a sparse Lidar scan at test time.}
\label{tab:sscbench-main}
\vspace{-3mm}
\end{table*}

%% file: figures/ssc_main.tex
\begin{figure*}
\centering
\captionsetup{type=figure}
\includegraphics[trim={0cm 0cm 0cm 0cm},clip,width=\linewidth]{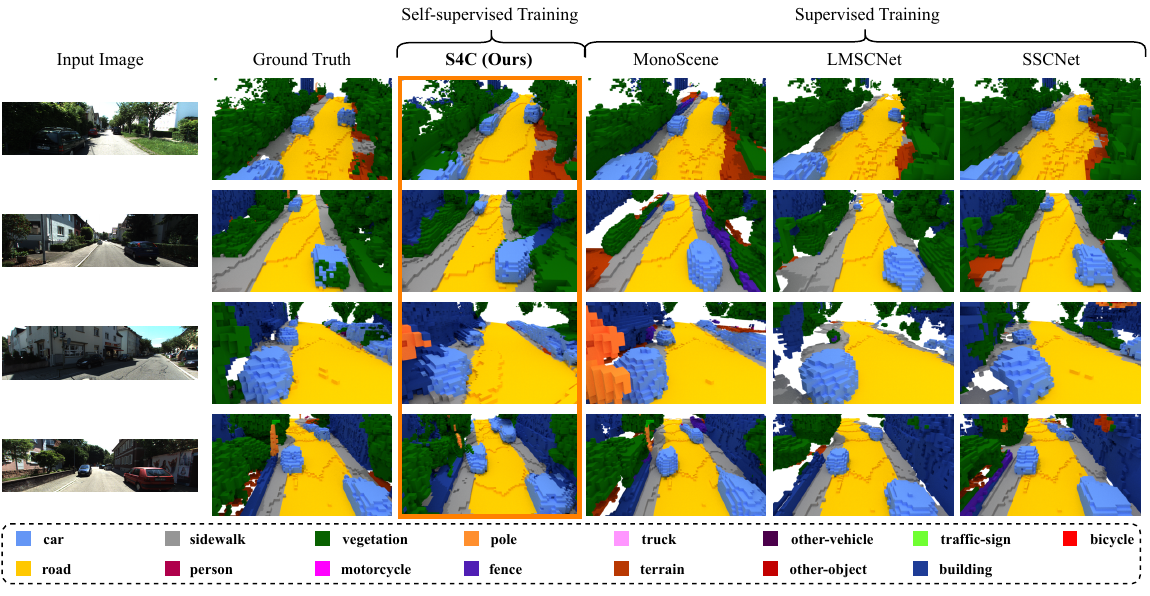}
\vspace{-0.7cm}
\captionof{figure}{
\textbf{Predicted voxel grids for SSCBench-KITTI-360}. The qualitative evaluation of our method on occupancy maps shows that our method is able to accurately reconstruct and label the scene. Especially a comparison to other image based methods like MonoScene shows, that \textbf{S4C} is able to recover details such as the driveway on the right in image 1. The resulting voxel occupancy from \textbf{S4C} shows fewer holes then for Lidar based training, which reproduce holes found in the ground truth. }
\label{fig:ssc_main}
\vspace{-0.2cm}
\end{figure*}

%% file: tables/sscbench_ablation.tex
\begin{table}[t]\centering
\renewcommand\tabcolsep{4pt}
\scriptsize
\begin{tabular}{ccc|ccc}
\toprule
Semantic & Photometric & View Offset & 12.8m & 25.6m & 51.2m \\
\midrule
\cmark & \xmark & 1s-4s & 31.11 & 32.04 & 27.88 \\
\xmark & \cmark & 1s-4s & 50.26 & 41.94 & 37.40 \\
\cmark & \cmark & 1s    & 52.00 & 41.96 & 36.67 \\
\textbf{\cmark} & \textbf{\cmark} & 1s-4s & \textbf{54.64} & \textbf{45.57} & \textbf{39.35} \\
\bottomrule
\end{tabular}
\caption{\textbf{Ablation studies on SSCBench-KITTI-360}. We report IoU for occupancy. The full model achieves the best performance.}
\label{tab:sscbench-ablation}
\vspace{-3mm}
\end{table}

%% file: figures/sscbench_ablation.tex
\begin{figure}
\centering
\captionsetup{type=figure}
\includegraphics[trim={0cm 0cm 0cm 0cm},clip,width=\linewidth]{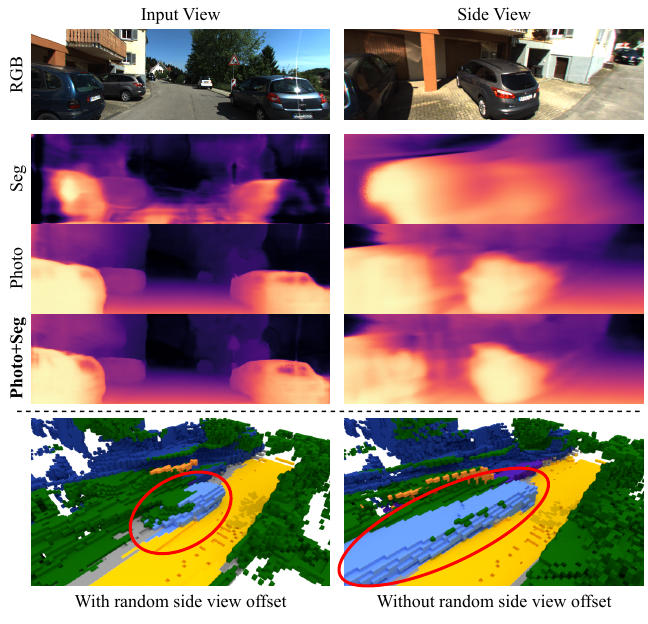}
\vspace{-0.7cm}
\captionof{figure}{
\textbf{Effects of different loss terms}.
We show expected ray termination depth for the input image and a corresponding side view for different loss configurations.
The full model produces sharpest results. We also show reconstruction with and without random offsetting the side views. This technique helps to correctly capture objects that are further away and reduce trailing effects.
}
\label{fig:sscbench-ablation}
\vspace{-0.2cm}
\end{figure}

%% file: tables/synth_seg.tex
\begin{table}[t]\centering
\renewcommand\tabcolsep{4pt}
\scriptsize
\begin{tabular}{l|c|ccc}
\toprule
Training Configuration &  KITTI GT & +5 & +10 & +15 \\
\midrule
Only Input Frame &  86.50\%  &  82.64\%  &  77.74\%  &  73.65\%  \\
Only Front    &  86.76\%  &  83.09\%  &  77.73\%  &  73.15\%  \\
\textbf{Full}          &  \textbf{87.81\%}  & \textbf{ 84.88\%}  &  \textbf{81.07\%}  &  \textbf{77.19\%}  \\
\bottomrule
\end{tabular}
\caption{\textbf{Accuracy of synthesized segmentation maps}. Segmentation maps from \textbf{S4C} produce accurate results when compared to the ground truth. Predicting segmentation maps for other frames (5, 10, and 15 timesteps in the future) shows the geometric accuracy of \textbf{S4C}. 
}
\label{tab:synth_seg}
\vspace{-3mm}
\end{table}

%% file: sec/5_discussion_and_conclusion.tex
\section{Conclusion}
\label{sec:conclusion}
This work presents \textbf{S4C}, a novel, image-based approach to semantic scene completion. 
It allows reconstructing both the 3D geometric structure and the semantic information of a scene from a single image. 
Although not trained on 3D ground truth, it achieves close to state-of-the-art performance on the KITTI-360 dataset within the SSCBench benchmark suite. 
As the first work on self-supervised semantic scene completion, \textbf{S4C} opens up the path towards scalable and cheap holistic 3D understanding.

%% file: sec/X_suppl.tex
\clearpage
\setcounter{page}{1}
\maketitlesupplementary

\section{Overview}

In this supplementary, we first in \cref{sec:sscbench} explain how we evaluated our methods on SSCBench \cite{li2023sscbench} and how we refined the invalid masks.
In \cref{sec:implementation_details}, we discuss further implementation details to enable easy reproduction of our experimental results.
\cref{sec:limitations} covers potential limitations of our method and \cref{sec:ethics} covers potential ethical concerns.
Finally, we provide the readers with more qualitative results in \cref{sec:add_results}.

\section{Video}
For further results, please watch our \textbf{provided video}, which contains per-frame predictions and comparisons for an entire subsequence of the KITTI-360 test set.

Comparisons with MonoScene \cite{cao2022monoscene}, SSCNet \cite{song2017semantic} and LMSCNet \cite{roldao2020lmscnet} happen at timestamps 00:16 and 02:21.

\section{SSCBench}
\label{sec:sscbench}

\subsection{Example of Training Data}
\input{figures/data_example}
\cref{fig:data_example} shows the eight frames we use to sample rays from per training sample. 
The different frames observe different parts of the scene.
This allows the network to learn geometry of the entire scene, even though areas are occluded in the input image.

\subsection{Refined Invalid Masks}
SSCBench processes annotated Lidar scans to produce labeled 3D voxel grids. 
If a voxel contains at least one Lidar measurement point, then it is considered occupied. The voxel label is obtained by majority voting over all contained Lidar measurement points.
However, only a small fraction of voxels in the scene contain Lidar points. 
To differentiate between empty and unknown (= ``invalid'') voxels, the authors perform raytracing from the Lidar origins.
Invalid voxels are ignored during evaluation.

During experiments, we observed that the unknown/empty voxel maps do not align perfectly with the voxel labels.
For example, a slice of one to two voxels below the street surface is considered known and empty, even though it is clearly underground and could not have been observed.

This perturbs evaluation results and gives methods that were trained directly on this ground truth a significant advantage.
They can learn to imitate this wrong behavior to achieve better accuracies, even though this does not reflect real-world accuracy.
To ensure a more level playing field, we therefore employ a simple strategy to refine the provided invalid masks.

All ground truth voxel grids have shape $\left(256 \times 256 \times 32\right)$ describing width, height, and depth (from a birds-eye perspective).
We consider every $z$ column separately.
If a voxel in a column has no valid occupied voxel below it (\ie it is most likely below ground or there exists no valid ground truth for the entire column), we consider it invalid.
As an additional check, we only consider voxels with $z \leq 7$. 
Through empirical study, we found that this corresponds to the $z$ level of the ground.
Note that \textbf{we never set occupied voxels to invalid}.
Per scene, only about 2-3\% percent of the voxels are affected by our invalid refinement strategy, as shown in \cref{fig:invalids_distribution}.
A mathematical precise formulation of this strategy is described in \cref{alg:invalids}.
\input{tables/additional_invalids}
\input{figures/invalids_distribution}
\input{figures/invalids_example}

As shown in \cref{fig:invalids_example}, our strategy finds many voxels below ground, that were previously considered to be empty and known, and sets them to invalid. 

For a fair comparison, we used the provided checkpoints and reran evaluation for all methods under the exact same setting as our method, using our invalid refinement strategy.
We observe that all methods achieve slightly better results than reported in SSCBench \cite{li2023sscbench}.
This affirms that the strategy is fair and provides a more accurate view on the performances of the different methods.

Both code and data for the invalid refinement strategy will be released upon acceptance of the paper to promote their use in future research and to facilitate fair comparisons.

\subsection{VoxFormer Reproduction}
Besides MonoScene \cite{cao2022monoscene}, another recently published work for supervised camera-based SSC is VoxFormer \cite{li2023voxformer}.
In the interest of a comprehensive evaluation, we wish to also provide benchmark results for this method.
However, despite using the official code base of VoxFormer and using the official checkpoint for KITTI-360 provided by SSCBench \cite{li2023sscbench}, we were not able to reproduce the benchmark results reported in \cite{li2023sscbench}.
Both when using the original SSCBench settings and when using our refined invalids, we achieve results that are not competitive with the other methods.

The results in the original SSCBench work suggest that VoxFormer is approximately on par with MonoScene on SSCBench-KITTI-360.
Therefore, in the interest of fairness, we choose to not report our VoxFormer results for now. 
We hope to work with the authors of VoxFormer and SSCBench to provide representative evaluation results for VoxFormer in the final version of this paper.

\section{Implementation Details}
\label{sec:implementation_details}

\input{tables/hyperparameters}
We use the same loss parameters for all trainings except for specific ablation studies.
A detailed overview of the exact hyperparameters is given in \cref{tab:hyperparam}.

Training on a single A40 GPU takes approximately 5 days until convergence.
We observed that our method is relatively robust when it comes to changes to hyperparameters.

\section{Limitations}
\label{sec:limitations}

While we believe that our proposed method achieves very strong results given that we use much weaker supervision compared to other approaches, it is not free of limitations.

As many other self-supervised methods for 3D reconstruction, we make the assumption that the world is made up of lambertian materials.
This means, that the color of a surface point does not change when observed from a different angle.
In our photometric loss, we therefore directly compare the pixel colors of different observations.
However, this assumption does not work for some categories. For example, our method struggles to correctly reconstruct windows of a car, that are both shiny and see-through and thus appear differently when viewed from a different angle.
Unlike methods that purely rely on photometric losses, however, our method is made more robust through the addition of the semantic loss, which does not suffer from this limitation.

Another limitation of our method is that our accuracy is dependent on the prediction quality an off-the-shelf image segmentation network.
While forward facing views from a driving vehicle are quite common in training datasets for segmentation networks, sideways facing views are rather rare.
Therefore, sometimes the segmentation predictions for the side views are not perfect.
The performance of our method could potentially be improved when using a better image segmentation network.

\section{Ethical and Safety Concerns}
\label{sec:ethics}

The datasets our method can be trained on potentially contain video footage of persons.
To ensure privacy, faces and other identifying features should be anonymized in the dataset.
Furthermore, it should be ensured that the dataset contains sufficient diversity.

Furthermore, training datatsets often only contain video material from a specific geographic location.
As traffic environments in different regions of the world or even a single country look vastly different, out method is not guaranteed to generalize to unknown environments.
Therefore, our method is not ready to be employed in safety-critical environments.

\section{Additonal Results}
\label{sec:add_results}

\subsection{Rendered 2D Segmentation Maps}
\input{figures/segmentations}
To extend Sec.\ 4.5 of the main paper, we show predicted segmentation masks for our model in \cref{fig:segmentations}.
Furthermore, we show synthesized segmentation masks for novel views.

%% file: figures/data_example.tex
\begin{figure*}
\centering
\captionsetup{type=figure}
\includegraphics[trim={0cm 0cm 0cm 0cm},clip,width=\linewidth]{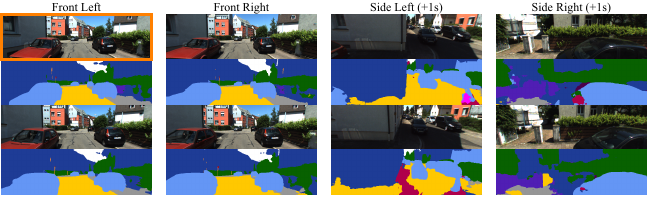}
\vspace{-0.7cm}
\captionof{figure}{
\textbf{Example of data sample}. Per training forward pass, we use eight frames (two timesteps, left / right, front / side) and corresponding segmentation maps which were predicted from an off-the-shelf segmentation network. 
The input image that is passed to the encoder-decoder is marked in orange.
The different frames observe different areas of the scene, allowing us to learn geometry and semantics even in occluded regions.}
\label{fig:data_example}
\vspace{-0.2cm}
\end{figure*}

%% file: tables/additional_invalids.tex
\begin{algorithm}
\caption{Refine invalid masks}\label{alg:invalids}
\begin{algorithmic}
\Require $\text{Inv} \in \{0, 1\}^{256\times 256\times 32}$
\Require $\text{Label} \in \{0, \dots, c\}^{256\times 256\times 32}$
    \State \Comment{Label $=0$ refers to known and empty.}

\State $\text{Inv}_\text{new} \gets 0^{256\times 256\times 32}$

\For{$x \in \operatorname{range}(256)$}
\For{$y \in \operatorname{range}(256)$}
\For{$z \in \operatorname{range}(7)$}
    \State \Comment{$z=7$ is street height.}
    \State $b \gets \bigwedge_{i = 0}^{z} \left(\text{Inv}[x, y, i] \lor \text{Label}[x, y, i] == 0\right)$
    \State $\text{Inv}_\text{new}[x, y, z] = b$
\EndFor
\EndFor
\EndFor
\State \Return $\text{Inv}_\text{new} \lor \text{Inv}$
\end{algorithmic}
\end{algorithm}

%% file: figures/invalids_distribution.tex
\begin{figure}
\centering
\captionsetup{type=figure}
\includegraphics[trim={0cm 0cm 0cm 0cm},clip,width=\linewidth]{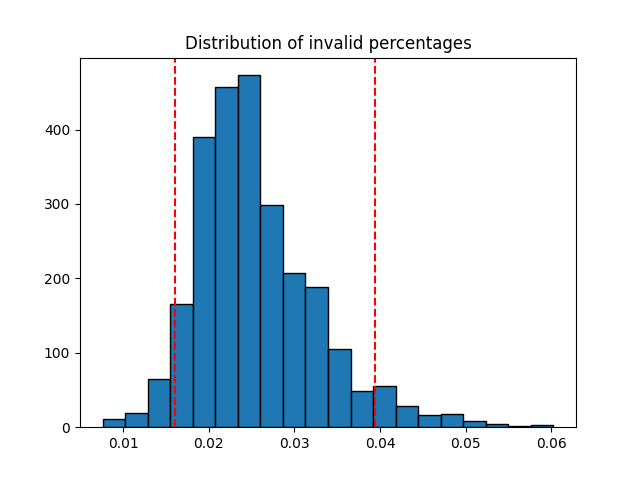}
\captionof{figure}{
\textbf{Percentage of voxels affected by our strategy.}
Per-scene, our invalid refinement strategy only affects approximately 2-3\% of the all voxels.
The histogram shows the percentage of affected voxels over the entire test set.
}
\label{fig:invalids_distribution}
\end{figure}

%% file: figures/invalids_example.tex
\begin{figure}
\centering
\captionsetup{type=figure}
\includegraphics[trim={0cm 0cm 0cm 0cm},clip,width=\linewidth]{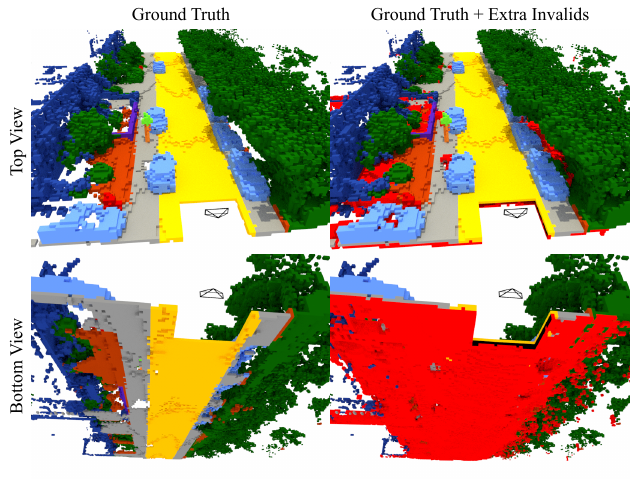}
\captionof{figure}{
\textbf{Example of refined invalids found by our strategy.}
We show the result of our invalid refinement strategy for an exemplary scene. The left column shows two views of the same scene with all occupied voxels. In the right column, we add all voxels (marked in bright red) that were previously considered empty and known, but identified by our strategy as invalid.
}
\label{fig:invalids_example}
\end{figure}

%% file: tables/hyperparameters.tex
\begin{table}[t]\centering
\renewcommand\tabcolsep{4pt}
\scriptsize
\begin{tabular}{lr|lr}
\toprule
$\lambda_\text{seg}$ & $0.02$     & Encoder-Decoder   & $34885032$ \\
$\lambda_\text{ph}$        & $1$        & $\phi_D$ (Density MLP)        & $6721$     \\
$\lambda_\text{eas}$          & $0.001$    & $\phi_S$ (Segmentation MLP) & $7891$     \\
$\eta$ (learning rate)       & $1e-04$ & Total weights             & $34899644$ \\
\midrule
$m$ (points per ray)   & $64$       & $z_\text{near}$           & $3$        \\
$n$ (number of frames)           & $8$        & $z_\text{far}$            & $80$      \\
\bottomrule
\end{tabular}
\caption{\textbf{Hyperparameters used during training}.
}
\label{tab:hyperparam}
\vspace{-3mm}
\end{table}

%% file: figures/segmentations.tex
\begin{figure*}
\centering
\captionsetup{type=figure}
\includegraphics[trim={0cm 0cm 0cm 0cm},clip,width=\linewidth]{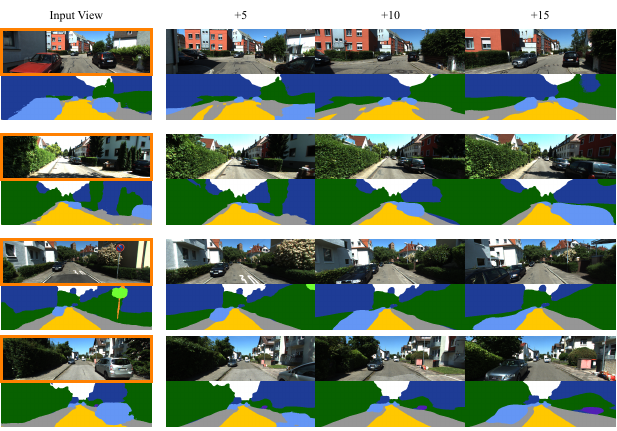}
\vspace{-0.7cm}
\captionof{figure}{
\textbf{Predicted segmentation masks by our model and semantic novel view synthesis}. For the a given input image (marked in orange), we can predict a dense segmentation mask. Additionally, our formulation allows to synthesize segmentation masks for later frames (+5, +10, +15) of the video. Even when the new view points are far away from the input view, the synthesized segmentation masks are clean and capture the semantics correctly.}
\label{fig:segmentations}
\vspace{-0.2cm}
\end{figure*}

%% file: main.bbl
\begin{thebibliography}{88}
\providecommand{\natexlab}[1]{#1}
\providecommand{\url}[1]{\texttt{#1}}
\expandafter\ifx\csname urlstyle\endcsname\relax
  \providecommand{\doi}[1]{doi: #1}\else
  \providecommand{\doi}{doi: \begingroup \urlstyle{rm}\Url}\fi

\bibitem[Aich et~al.(2021)Aich, Vianney, Islam, and Liu]{aich2021bidirectional}
Shubhra Aich, Jean Marie~Uwabeza Vianney, Md~Amirul Islam, and Mannat
  Kaur~Bingbing Liu.
\newblock Bidirectional attention network for monocular depth estimation.
\newblock In \emph{IEEE International Conference on Robotics and Automation},
  2021.

\bibitem[Alnaggar et~al.(2021)Alnaggar, Afifi, Amer, and
  ElHelw]{alnaggar2021multi}
Yara~Ali Alnaggar, Mohamed Afifi, Karim Amer, and Mohamed ElHelw.
\newblock Multi projection fusion for real-time semantic segmentation of 3d
  lidar point clouds.
\newblock In \emph{Proceedings of the IEEE/CVF winter conference on
  applications of computer vision}, 2021.

\bibitem[Armeni et~al.(2019)Armeni, He, Gwak, Zamir, Fischer, Malik, and
  Savarese]{armeni20193d}
Iro Armeni, Zhi-Yang He, JunYoung Gwak, Amir~R Zamir, Martin Fischer, Jitendra
  Malik, and Silvio Savarese.
\newblock 3d scene graph: A structure for unified semantics, 3d space, and
  camera.
\newblock In \emph{CVPR}, 2019.

\bibitem[Behley et~al.(2019)Behley, Garbade, Milioto, Quenzel, Behnke,
  Stachniss, and Gall]{behley2019semantickitti}
Jens Behley, Martin Garbade, Andres Milioto, Jan Quenzel, Sven Behnke, Cyrill
  Stachniss, and Jurgen Gall.
\newblock Semantickitti: A dataset for semantic scene understanding of lidar
  sequences.
\newblock In \emph{Proceedings of the IEEE/CVF International Conference on
  Computer Vision}, pages 9297--9307, 2019.

\bibitem[Beltr{\'a}n et~al.(2018)Beltr{\'a}n, Guindel, Moreno, Cruzado, Garcia,
  and De~La~Escalera]{beltran2018birdnet}
Jorge Beltr{\'a}n, Carlos Guindel, Francisco~Miguel Moreno, Daniel Cruzado,
  Fernando Garcia, and Arturo De~La~Escalera.
\newblock Birdnet: a 3d object detection framework from lidar information.
\newblock In \emph{2018 21st International Conference on Intelligent
  Transportation Systems (ITSC)}. IEEE, 2018.

\bibitem[Caesar et~al.(2020)Caesar, Bankiti, Lang, Vora, Liong, Xu, Krishnan,
  Pan, Baldan, and Beijbom]{caesar2020nuscenes}
Holger Caesar, Varun Bankiti, Alex~H Lang, Sourabh Vora, Venice~Erin Liong,
  Qiang Xu, Anush Krishnan, Yu Pan, Giancarlo Baldan, and Oscar Beijbom.
\newblock nuscenes: A multimodal dataset for autonomous driving.
\newblock In \emph{Proceedings of the IEEE/CVF conference on computer vision
  and pattern recognition}, pages 11621--11631, 2020.

\bibitem[Cai et~al.(2021)Cai, Chen, Zhang, Lin, Wang, and Li]{cai2021semantic}
Yingjie Cai, Xuesong Chen, Chao Zhang, Kwan-Yee Lin, Xiaogang Wang, and
  Hongsheng Li.
\newblock Semantic scene completion via integrating instances and scene
  in-the-loop.
\newblock In \emph{Proceedings of the IEEE/CVF Conference on Computer Vision
  and Pattern Recognition}, pages 324--333, 2021.

\bibitem[Cao and de~Charette(2022)]{cao2022monoscene}
Anh-Quan Cao and Raoul de Charette.
\newblock Monoscene: Monocular 3d semantic scene completion.
\newblock In \emph{CVPR}, 2022.

\bibitem[Cao and de~Charette(2023)]{cao2023scenerf}
Anh-Quan Cao and Raoul de Charette.
\newblock Scenerf: Self-supervised monocular 3d scene reconstruction with
  radiance fields.
\newblock In \emph{ICCV}, 2023.

\bibitem[Chen et~al.(2020)Chen, Lin, Qian, Zeng, and Li]{chen20203d}
Xiaokang Chen, Kwan-Yee Lin, Chen Qian, Gang Zeng, and Hongsheng Li.
\newblock 3d sketch-aware semantic scene completion via semi-supervised
  structure prior.
\newblock In \emph{Proceedings of the IEEE/CVF Conference on Computer Vision
  and Pattern Recognition}, pages 4193--4202, 2020.

\bibitem[Cheng et~al.(2020)Cheng, Collins, Zhu, Liu, Huang, Adam, and
  Chen]{cheng2020panoptic}
Bowen Cheng, Maxwell~D Collins, Yukun Zhu, Ting Liu, Thomas~S Huang, Hartwig
  Adam, and Liang-Chieh Chen.
\newblock Panoptic-deeplab: A simple, strong, and fast baseline for bottom-up
  panoptic segmentation.
\newblock In \emph{Proceedings of the IEEE/CVF conference on computer vision
  and pattern recognition}, pages 12475--12485, 2020.

\bibitem[Cheng et~al.(2021)Cheng, Agia, Ren, Li, and Bingbing]{cheng2021s3cnet}
Ran Cheng, Christopher Agia, Yuan Ren, Xinhai Li, and Liu Bingbing.
\newblock S3cnet: A sparse semantic scene completion network for lidar point
  clouds.
\newblock In \emph{Conference on Robot Learning}, pages 2148--2161. PMLR, 2021.

\bibitem[Choy et~al.(2016)Choy, Xu, Gwak, Chen, and Savarese]{choy20163d}
Christopher~B Choy, Danfei Xu, JunYoung Gwak, Kevin Chen, and Silvio Savarese.
\newblock 3d-r2n2: A unified approach for single and multi-view 3d object
  reconstruction.
\newblock In \emph{European conference on computer vision}, pages 628--644.
  Springer, 2016.

\bibitem[Cordts et~al.(2016)Cordts, Omran, Ramos, Rehfeld, Enzweiler, Benenson,
  Franke, Roth, and Schiele]{cordts2016cityscapes}
Marius Cordts, Mohamed Omran, Sebastian Ramos, Timo Rehfeld, Markus Enzweiler,
  Rodrigo Benenson, Uwe Franke, Stefan Roth, and Bernt Schiele.
\newblock The cityscapes dataset for semantic urban scene understanding.
\newblock In \emph{Proceedings of the IEEE conference on computer vision and
  pattern recognition}, pages 3213--3223, 2016.

\bibitem[Eigen et~al.(2014)Eigen, Puhrsch, and Fergus]{eigen2014depth}
David Eigen, Christian Puhrsch, and Rob Fergus.
\newblock Depth map prediction from a single image using a multi-scale deep
  network.
\newblock \emph{NeurIPS}, 27, 2014.

\bibitem[Fan et~al.(2017)Fan, Su, and Guibas]{fan2017point}
Haoqiang Fan, Hao Su, and Leonidas~J Guibas.
\newblock A point set generation network for 3d object reconstruction from a
  single image.
\newblock In \emph{Proceedings of the IEEE conference on computer vision and
  pattern recognition}, pages 605--613, 2017.

\bibitem[Fu et~al.(2018)Fu, Gong, Wang, Batmanghelich, and Tao]{fu2018deep}
Huan Fu, Mingming Gong, Chaohui Wang, Kayhan Batmanghelich, and Dacheng Tao.
\newblock Deep ordinal regression network for monocular depth estimation.
\newblock In \emph{CVPR}, 2018.

\bibitem[Fu et~al.(2022)Fu, Zhang, Chen, Lu, Zhu, Zhou, Geiger, and
  Liao]{fu2022panoptic}
Xiao Fu, Shangzhan Zhang, Tianrun Chen, Yichong Lu, Lanyun Zhu, Xiaowei Zhou,
  Andreas Geiger, and Yiyi Liao.
\newblock Panoptic nerf: 3d-to-2d label transfer for panoptic urban scene
  segmentation.
\newblock In \emph{International Conference on 3D Vision (3DV)}, 2022.

\bibitem[Godard et~al.(2017)Godard, Mac~Aodha, and
  Brostow]{godard2017unsupervised}
Cl{\'e}ment Godard, Oisin Mac~Aodha, and Gabriel~J Brostow.
\newblock Unsupervised monocular depth estimation with left-right consistency.
\newblock In \emph{ICCV}, pages 270--279, 2017.

\bibitem[Godard et~al.(2019)Godard, Mac~Aodha, Firman, and
  Brostow]{godard2019digging}
Cl{\'e}ment Godard, Oisin Mac~Aodha, Michael Firman, and Gabriel~J Brostow.
\newblock Digging into self-supervised monocular depth estimation.
\newblock In \emph{ICCV}, pages 3828--3838, 2019.

\bibitem[Gupta et~al.(2013)Gupta, Arbelaez, and Malik]{gupta2013perceptual}
Saurabh Gupta, Pablo Arbelaez, and Jitendra Malik.
\newblock Perceptual organization and recognition of indoor scenes from rgb-d
  images.
\newblock In \emph{Proceedings of the IEEE conference on computer vision and
  pattern recognition}, pages 564--571, 2013.

\bibitem[Han et~al.(2021)Han, Laga, and Bennamoun]{han2021sota_overview}
Xian-Feng Han, Hamid Laga, and Mohammed Bennamoun.
\newblock Image-based 3d object reconstruction: State-of-the-art and trends in
  the deep learning era.
\newblock \emph{IEEE TPAMI}, page 1578—1604, 2021.

\bibitem[Hartley and Zisserman(2003)]{hartley2003multiple}
Richard Hartley and Andrew Zisserman.
\newblock \emph{Multiple view geometry in computer vision}.
\newblock Cambridge university press, 2003.

\bibitem[He et~al.(2016)He, Zhang, Ren, and Sun]{he2016deep}
Kaiming He, Xiangyu Zhang, Shaoqing Ren, and Jian Sun.
\newblock Deep residual learning for image recognition.
\newblock In \emph{Proceedings of the IEEE conference on computer vision and
  pattern recognition}, pages 770--778, 2016.

\bibitem[He et~al.(2019)He, Huang, Yi, Zhou, Wu, Wang, and
  Soatto]{he2019geonet}
Tong He, Haibin Huang, Li Yi, Yuqian Zhou, Chihao Wu, Jue Wang, and Stefano
  Soatto.
\newblock Geonet: Deep geodesic networks for point cloud analysis.
\newblock In \emph{CVPR}, 2019.

\bibitem[Hermans et~al.(2014)Hermans, Floros, and Leibe]{hermans2014dense}
Alexander Hermans, Georgios Floros, and Bastian Leibe.
\newblock Dense 3d semantic mapping of indoor scenes from rgb-d images.
\newblock In \emph{IEEE International Conference on Robotics and Automation},
  2014.

\bibitem[Huang et~al.(2023)Huang, Zheng, Zhang, Zhou, and Lu]{huang2023tri}
Yuanhui Huang, Wenzhao Zheng, Yunpeng Zhang, Jie Zhou, and Jiwen Lu.
\newblock Tri-perspective view for vision-based 3d semantic occupancy
  prediction.
\newblock In \emph{Proceedings of the IEEE/CVF Conference on Computer Vision
  and Pattern Recognition}, pages 9223--9232, 2023.

\bibitem[Jatavallabhula et~al.(2023)Jatavallabhula, Kuwajerwala, Gu, Omama,
  Chen, Li, Iyer, Saryazdi, Keetha, Tewari, Tenenbaum, {de Melo}, Krishna,
  Paull, Shkurti, and Torralba]{jatavallabhula2023conceptfusion}
{Krishna Murthy} Jatavallabhula, Alihusein Kuwajerwala, Qiao Gu, Mohd Omama,
  Tao Chen, Shuang Li, Ganesh Iyer, Soroush Saryazdi, Nikhil Keetha, Ayush
  Tewari, {Joshua B.} Tenenbaum, {Celso Miguel} {de Melo}, Madhava Krishna,
  Liam Paull, Florian Shkurti, and Antonio Torralba.
\newblock Conceptfusion: Open-set multimodal 3d mapping.
\newblock \emph{arXiv}, 2023.

\bibitem[Kajiya and Von~Herzen(1984)]{kajiya1984ray}
James~T Kajiya and Brian~P Von~Herzen.
\newblock Ray tracing volume densities.
\newblock \emph{ACM SIGGRAPH computer graphics}, 18\penalty0 (3):\penalty0
  165--174, 1984.

\bibitem[Kundu et~al.(2020)Kundu, Yin, Fathi, Ross, Brewington, Funkhouser, and
  Pantofaru]{kundu2020virtual}
Abhijit Kundu, Xiaoqi Yin, Alireza Fathi, David Ross, Brian Brewington, Thomas
  Funkhouser, and Caroline Pantofaru.
\newblock Virtual multi-view fusion for 3d semantic segmentation.
\newblock In \emph{ECCV}, 2020.

\bibitem[Kundu et~al.(2022)Kundu, Genova, Yin, Fathi, Pantofaru, Guibas,
  Tagliasacchi, Dellaert, and Funkhouser]{KunduCVPR2022PNF}
Abhijit Kundu, Kyle Genova, Xiaoqi Yin, Alireza Fathi, Caroline Pantofaru,
  Leonidas Guibas, Andrea Tagliasacchi, Frank Dellaert, and Thomas Funkhouser.
\newblock {Panoptic Neural Fields: A Semantic Object-Aware Neural Scene
  Representation}.
\newblock In \emph{CVPR}, 2022.

\bibitem[Kuznietsov et~al.(2017)Kuznietsov, Stuckler, and
  Leibe]{kuznietsov2017semi}
Yevhen Kuznietsov, Jorg Stuckler, and Bastian Leibe.
\newblock Semi-supervised deep learning for monocular depth map prediction.
\newblock In \emph{CVPR}, 2017.

\bibitem[Lee et~al.(2021)Lee, Lee, Kim, Yi, and Kim]{lee2021patch}
Sihaeng Lee, Janghyeon Lee, Byungju Kim, Eojindl Yi, and Junmo Kim.
\newblock Patch-wise attention network for monocular depth estimation.
\newblock In \emph{AAAI}, 2021.

\bibitem[Li et~al.(2019{\natexlab{a}})Li, Liu, Gong, Shi, Yuan, Zhao, and
  Reid]{li2019rgbd}
Jie Li, Yu Liu, Dong Gong, Qinfeng Shi, Xia Yuan, Chunxia Zhao, and Ian Reid.
\newblock Rgbd based dimensional decomposition residual network for 3d semantic
  scene completion.
\newblock In \emph{Proceedings of the IEEE/CVF Conference on Computer Vision
  and Pattern Recognition}, pages 7693--7702, 2019{\natexlab{a}}.

\bibitem[Li et~al.(2019{\natexlab{b}})Li, Liu, Yuan, Zhao, Siegwart, Reid, and
  Cadena]{li2019depth}
Jie Li, Yu Liu, Xia Yuan, Chunxia Zhao, Roland Siegwart, Ian Reid, and Cesar
  Cadena.
\newblock Depth based semantic scene completion with position importance aware
  loss.
\newblock \emph{IEEE Robotics and Automation Letters}, 5\penalty0 (1):\penalty0
  219--226, 2019{\natexlab{b}}.

\bibitem[Li et~al.(2020)Li, Han, Wang, Liu, and Yuan]{li2020anisotropic}
Jie Li, Kai Han, Peng Wang, Yu Liu, and Xia Yuan.
\newblock Anisotropic convolutional networks for 3d semantic scene completion.
\newblock In \emph{Proceedings of the IEEE/CVF Conference on Computer Vision
  and Pattern Recognition}, pages 3351--3359, 2020.

\bibitem[Li et~al.(2021)Li, Shi, Liu, Zhao, Zhou, and
  Zhang]{li2021semisupervised}
Pengfei Li, Yongliang Shi, Tianyu Liu, Hao Zhao, Guyue Zhou, and Ya-Qin Zhang.
\newblock Semi-supervised implicit scene completion from sparse lidar, 2021.

\bibitem[Li et~al.(2023{\natexlab{a}})Li, Li, Liu, Gong, Li, Chen, Wang, Li,
  Jiang, Yu, Wang, Zhao, Yu, and Feng]{li2023sscbench}
Yiming Li, Sihang Li, Xinhao Liu, Moonjun Gong, Kenan Li, Nuo Chen, Zijun Wang,
  Zhiheng Li, Tao Jiang, Fisher Yu, Yue Wang, Hang Zhao, Zhiding Yu, and Chen
  Feng.
\newblock Sscbench: A large-scale 3d semantic scene completion benchmark for
  autonomous driving.
\newblock \emph{arXiv preprint arXiv:2306.09001}, 2023{\natexlab{a}}.

\bibitem[Li et~al.(2023{\natexlab{b}})Li, Yu, Choy, Xiao, Alvarez, Fidler,
  Feng, and Anandkumar]{li2023voxformer}
Yiming Li, Zhiding Yu, Christopher Choy, Chaowei Xiao, Jose~M Alvarez, Sanja
  Fidler, Chen Feng, and Anima Anandkumar.
\newblock Voxformer: Sparse voxel transformer for camera-based 3d semantic
  scene completion.
\newblock In \emph{CVPR}, 2023{\natexlab{b}}.

\bibitem[Li et~al.(2022{\natexlab{a}})Li, Chen, Liu, and
  Jiang]{li2022depthformer}
Zhenyu Li, Zehui Chen, Xianming Liu, and Junjun Jiang.
\newblock Depthformer: Exploiting long-range correlation and local information
  for accurate monocular depth estimation.
\newblock \emph{arXiv preprint arXiv:2203.14211}, 2022{\natexlab{a}}.

\bibitem[Li et~al.(2022{\natexlab{b}})Li, Wang, Liu, and
  Jiang]{li2022binsformer}
Zhenyu Li, Xuyang Wang, Xianming Liu, and Junjun Jiang.
\newblock Binsformer: Revisiting adaptive bins for monocular depth estimation.
\newblock \emph{arXiv preprint arXiv:2204.00987}, 2022{\natexlab{b}}.

\bibitem[Li et~al.(2023{\natexlab{c}})Li, Yu, Austin, Fang, Lan, Kautz, and
  Alvarez]{li2023fb}
Zhiqi Li, Zhiding Yu, David Austin, Mingsheng Fang, Shiyi Lan, Jan Kautz, and
  Jose~M Alvarez.
\newblock Fb-occ: 3d occupancy prediction based on forward-backward view
  transformation.
\newblock \emph{arXiv preprint arXiv:2307.01492}, 2023{\natexlab{c}}.

\bibitem[Liao et~al.(2022)Liao, Xie, and Geiger]{liao2022kitti}
Yiyi Liao, Jun Xie, and Andreas Geiger.
\newblock Kitti-360: A novel dataset and benchmarks for urban scene
  understanding in 2d and 3d.
\newblock \emph{IEEE Transactions on Pattern Analysis and Machine
  Intelligence}, 2022.

\bibitem[Liu et~al.(2015)Liu, Shen, Lin, and Reid]{liu2015learning}
Fayao Liu, Chunhua Shen, Guosheng Lin, and Ian Reid.
\newblock Learning depth from single monocular images using deep convolutional
  neural fields.
\newblock \emph{IEEE TPAMI}, 2015.

\bibitem[Liu et~al.(2018)Liu, Hu, Zeng, Tang, Jin, Han, and Li]{liu2018see}
Shice Liu, Yu Hu, Yiming Zeng, Qiankun Tang, Beibei Jin, Yinhe Han, and Xiaowei
  Li.
\newblock See and think: Disentangling semantic scene completion.
\newblock In \emph{Advances in Neural Information Processing Systems}, 2018.

\bibitem[Lyu et~al.(2021)Lyu, Liu, Wang, Kong, Liu, Liu, Chen, and
  Yuan]{lyu2021hr}
Xiaoyang Lyu, Liang Liu, Mengmeng Wang, Xin Kong, Lina Liu, Yong Liu, Xinxin
  Chen, and Yi Yuan.
\newblock Hr-depth: High resolution self-supervised monocular depth estimation.
\newblock In \emph{AAAI}, pages 2294--2301, 2021.

\bibitem[Mascaro et~al.(2021)Mascaro, Teixeira, and Chli]{mascaro2021diffuser}
Ruben Mascaro, Lucas Teixeira, and Margarita Chli.
\newblock Diffuser: Multi-view 2d-to-3d label diffusion for semantic scene
  segmentation.
\newblock In \emph{IEEE International Conference on Robotics and Automation},
  2021.

\bibitem[Maturana and Scherer(2015)]{maturana2015voxnet}
Daniel Maturana and Sebastian Scherer.
\newblock Voxnet: A 3d convolutional neural network for real-time object
  recognition.
\newblock In \emph{2015 IEEE/RSJ international conference on intelligent robots
  and systems (IROS)}. IEEE, 2015.

\bibitem[Max(1995)]{max1995optical}
Nelson Max.
\newblock Optical models for direct volume rendering.
\newblock \emph{IEEE Transactions on Visualization and Computer Graphics},
  1\penalty0 (2):\penalty0 99--108, 1995.

\bibitem[McCormac et~al.(2017)McCormac, Handa, Davison, and
  Leutenegger]{mccormac2017semanticfusion}
John McCormac, Ankur Handa, Andrew Davison, and Stefan Leutenegger.
\newblock Semanticfusion: Dense 3d semantic mapping with convolutional neural
  networks.
\newblock In \emph{IEEE International Conference on Robotics and Automation},
  2017.

\bibitem[Miao et~al.(2023)Miao, Liu, Chen, Gong, Xu, Hu, and
  Zhou]{miao2023occdepth}
Ruihang Miao, Weizhou Liu, Mingrui Chen, Zheng Gong, Weixin Xu, Chen Hu, and
  Shuchang Zhou.
\newblock Occdepth: A depth-aware method for 3d semantic scene completion.
\newblock \emph{arXiv preprint arXiv:2302.13540}, 2023.

\bibitem[Mildenhall et~al.(2020)Mildenhall, Srinivasan, Tancik, Barron,
  Ramamoorthi, and Ng]{mildenhall2020nerf}
Ben Mildenhall, Pratul~P. Srinivasan, Matthew Tancik, Jonathan~T. Barron, Ravi
  Ramamoorthi, and Ren Ng.
\newblock Nerf: Representing scenes as neural radiance fields for view
  synthesis.
\newblock In \emph{ECCV}, 2020.

\bibitem[Milioto et~al.(2019)Milioto, Vizzo, Behley, and
  Stachniss]{milioto2019rangenet++}
Andres Milioto, Ignacio Vizzo, Jens Behley, and Cyrill Stachniss.
\newblock Rangenet++: Fast and accurate lidar semantic segmentation.
\newblock In \emph{2019 IEEE/RSJ international conference on intelligent robots
  and systems (IROS)}. IEEE, 2019.

\bibitem[Paszke et~al.(2019)Paszke, Gross, Massa, Lerer, Bradbury, Chanan,
  Killeen, Lin, Gimelshein, Antiga, et~al.]{paszke2019pytorch}
Adam Paszke, Sam Gross, Francisco Massa, Adam Lerer, James Bradbury, Gregory
  Chanan, Trevor Killeen, Zeming Lin, Natalia Gimelshein, Luca Antiga, et~al.
\newblock Pytorch: An imperative style, high-performance deep learning library.
\newblock \emph{Advances in neural information processing systems}, 32, 2019.

\bibitem[Peng et~al.(2023)Peng, Genova, Jiang, Tagliasacchi, Pollefeys, and
  Funkhouser]{peng2023openscene}
Songyou Peng, Kyle Genova, Chiyu~"Max" Jiang, Andrea Tagliasacchi, Marc
  Pollefeys, and Thomas Funkhouser.
\newblock Openscene: 3d scene understanding with open vocabularies.
\newblock In \emph{CVPR}, 2023.

\bibitem[Qi et~al.(2016)Qi, Su, Nie{\ss}ner, Dai, Yan, and
  Guibas]{qi2016volumetric}
Charles~R Qi, Hao Su, Matthias Nie{\ss}ner, Angela Dai, Mengyuan Yan, and
  Leonidas~J Guibas.
\newblock Volumetric and multi-view cnns for object classification on 3d data.
\newblock In \emph{CVPR}, 2016.

\bibitem[Qi et~al.(2017{\natexlab{a}})Qi, Su, Mo, and Guibas]{qi2017pointnet}
Charles~R Qi, Hao Su, Kaichun Mo, and Leonidas~J Guibas.
\newblock Pointnet: Deep learning on point sets for 3d classification and
  segmentation.
\newblock In \emph{CVPR}, 2017{\natexlab{a}}.

\bibitem[Qi et~al.(2017{\natexlab{b}})Qi, Yi, Su, and Guibas]{qi2017pointnet++}
Charles~Ruizhongtai Qi, Li Yi, Hao Su, and Leonidas~J Guibas.
\newblock Pointnet++: Deep hierarchical feature learning on point sets in a
  metric space.
\newblock \emph{Advances in neural information processing systems},
  2017{\natexlab{b}}.

\bibitem[Rist et~al.(2021)Rist, Emmerichs, Enzweiler, and
  Gavrila]{rist2021semantic}
Christoph~B Rist, David Emmerichs, Markus Enzweiler, and Dariu~M Gavrila.
\newblock Semantic scene completion using local deep implicit functions on
  lidar data.
\newblock \emph{IEEE transactions on pattern analysis and machine
  intelligence}, 44\penalty0 (10):\penalty0 7205--7218, 2021.

\bibitem[Roldao et~al.(2020)Roldao, de~Charette, and
  Verroust-Blondet]{roldao2020lmscnet}
Luis Roldao, Raoul de Charette, and Anne Verroust-Blondet.
\newblock Lmscnet: Lightweight multiscale 3d semantic completion.
\newblock In \emph{2020 International Conference on 3D Vision (3DV)}, 2020.

\bibitem[Roldao et~al.(2022)Roldao, De~Charette, and
  Verroust-Blondet]{roldao20223d}
Luis Roldao, Raoul De~Charette, and Anne Verroust-Blondet.
\newblock 3d semantic scene completion: A survey.
\newblock \emph{IJCV}, 2022.

\bibitem[Siddiqui et~al.(2023)Siddiqui, Porzi, Bul{\`o}, M{\"u}ller,
  Nie{\ss}ner, Dai, and Kontschieder]{siddiqui2023panoptic}
Yawar Siddiqui, Lorenzo Porzi, Samuel~Rota Bul{\`o}, Norman M{\"u}ller,
  Matthias Nie{\ss}ner, Angela Dai, and Peter Kontschieder.
\newblock Panoptic lifting for 3d scene understanding with neural fields.
\newblock In \emph{CVPR}, 2023.

\bibitem[Sima et~al.(2023)Sima, Tong, Wang, Chen, Wu, Deng, Gu, Lu, Luo, Lin,
  and Li]{sima2023_occnet}
Chonghao Sima, Wenwen Tong, Tai Wang, Li Chen, Silei Wu, Hanming Deng, Yi Gu,
  Lewei Lu, Ping Luo, Dahua Lin, and Hongyang Li.
\newblock Scene as occupancy.
\newblock \emph{arXiv preprint arXiv:2306.02851}, 2023.

\bibitem[Song et~al.(2017)Song, Yu, Zeng, Chang, Savva, and
  Funkhouser]{song2017semantic}
Shuran Song, Fisher Yu, Andy Zeng, Angel~X Chang, Manolis Savva, and Thomas
  Funkhouser.
\newblock Semantic scene completion from a single depth image.
\newblock In \emph{CVPR}, 2017.

\bibitem[Spencer et~al.(2020)Spencer, Bowden, and Hadfield]{spencer2020defeat}
Jaime Spencer, Richard Bowden, and Simon Hadfield.
\newblock Defeat-net: General monocular depth via simultaneous unsupervised
  representation learning.
\newblock In \emph{CVPR}, pages 14402--14413, 2020.

\bibitem[Sun et~al.(2020)Sun, Kretzschmar, Dotiwalla, Chouard, Patnaik, Tsui,
  Guo, Zhou, Chai, Caine, Vasudevan, Han, Ngiam, Zhao, Timofeev, Ettinger,
  Krivokon, Gao, Joshi, Zhang, Shlens, Chen, and Anguelov]{sun2020scalability}
Pei Sun, Henrik Kretzschmar, Xerxes Dotiwalla, Aurelien Chouard, Vijaysai
  Patnaik, Paul Tsui, James Guo, Yin Zhou, Yuning Chai, Benjamin Caine, Vijay
  Vasudevan, Wei Han, Jiquan Ngiam, Hang Zhao, Aleksei Timofeev, Scott
  Ettinger, Maxim Krivokon, Amy Gao, Aditya Joshi, Yu Zhang, Jonathon Shlens,
  Zhifeng Chen, and Dragomir Anguelov.
\newblock Scalability in perception for autonomous driving: Waymo open dataset.
\newblock In \emph{CVPR}, 2020.

\bibitem[Thrun and Wegbreit(2005)]{thrun2005shape}
S. Thrun and B. Wegbreit.
\newblock Shape from symmetry.
\newblock In \emph{Tenth IEEE International Conference on Computer Vision
  (ICCV'05) Volume 1}, pages 1824--1831 Vol. 2, 2005.

\bibitem[Tulsiani et~al.(2017)Tulsiani, Zhou, Efros, and
  Malik]{tulsiani2017multi}
Shubham Tulsiani, Tinghui Zhou, Alexei~A Efros, and Jitendra Malik.
\newblock Multi-view supervision for single-view reconstruction via
  differentiable ray consistency.
\newblock In \emph{Proceedings of the IEEE conference on computer vision and
  pattern recognition}, pages 2626--2634, 2017.

\bibitem[Vora et~al.(2021)Vora, Radwan, Greff, Meyer, Genova, Sajjadi, Pot,
  Tagliasacchi, and Duckworth]{vora2021nesf}
Suhani Vora, Noha Radwan, Klaus Greff, Henning Meyer, Kyle Genova, Mehdi S.~M.
  Sajjadi, Etienne Pot, Andrea Tagliasacchi, and Daniel Duckworth.
\newblock Nesf: Neural semantic fields for generalizable semantic segmentation
  of 3d scenes, 2021.

\bibitem[Wang et~al.(2022)Wang, Chen, and Yang]{wang2022dm}
Bing Wang, Lu Chen, and Bo Yang.
\newblock Dm-nerf: 3d scene geometry decomposition and manipulation from 2d
  images.
\newblock \emph{arXiv preprint arXiv:2208.07227}, 2022.

\bibitem[Wang et~al.(2019{\natexlab{a}})Wang, Cheng, Sohel, Bennamoun, and
  Li]{wang2019normalnet}
Cheng Wang, Ming Cheng, Ferdous Sohel, Mohammed Bennamoun, and Jonathan Li.
\newblock Normalnet: A voxel-based cnn for 3d object classification and
  retrieval.
\newblock \emph{Neurocomputing}, 2019{\natexlab{a}}.

\bibitem[Wang et~al.(2019{\natexlab{b}})Wang, Sun, Liu, Sarma, Bronstein, and
  Solomon]{wang2019dynamic}
Yue Wang, Yongbin Sun, Ziwei Liu, Sanjay~E Sarma, Michael~M Bronstein, and
  Justin~M Solomon.
\newblock Dynamic graph cnn for learning on point clouds.
\newblock \emph{ACM Transactions on Graphics (tog)}, 2019{\natexlab{b}}.

\bibitem[Wang et~al.(2004)Wang, Bovik, Sheikh, and Simoncelli]{wang2004image}
Zhou Wang, Alan~C Bovik, Hamid~R Sheikh, and Eero~P Simoncelli.
\newblock Image quality assessment: from error visibility to structural
  similarity.
\newblock \emph{IEEE transactions on image processing}, 13\penalty0
  (4):\penalty0 600--612, 2004.

\bibitem[Wimbauer et~al.(2021)Wimbauer, Yang, Von~Stumberg, Zeller, and
  Cremers]{wimbauer2021monorec}
Felix Wimbauer, Nan Yang, Lukas Von~Stumberg, Niclas Zeller, and Daniel
  Cremers.
\newblock Monorec: Semi-supervised dense reconstruction in dynamic environments
  from a single moving camera.
\newblock In \emph{CVPR}, 2021.

\bibitem[Wimbauer et~al.(2023)Wimbauer, Yang, Rupprecht, and
  Cremers]{wimbauer2023behind}
Felix Wimbauer, Nan Yang, Christian Rupprecht, and Daniel Cremers.
\newblock Behind the scenes: Density fields for single view reconstruction.
\newblock In \emph{CVPR}, 2023.

\bibitem[Wu et~al.(2018)Wu, Wan, Yue, and Keutzer]{wu2018squeezeseg}
Bichen Wu, Alvin Wan, Xiangyu Yue, and Kurt Keutzer.
\newblock Squeezeseg: Convolutional neural nets with recurrent crf for
  real-time road-object segmentation from 3d lidar point cloud.
\newblock In \emph{IEEE International Conference on Robotics and Automation}.
  IEEE, 2018.

\bibitem[Yan et~al.(2016)Yan, Yang, Yumer, Guo, and Lee]{yan2016perspective}
Xinchen Yan, Jimei Yang, Ersin Yumer, Yijie Guo, and Honglak Lee.
\newblock Perspective transformer nets: Learning single-view 3d object
  reconstruction without 3d supervision.
\newblock In \emph{Advances in neural information processing systems}, 2016.

\bibitem[Yan et~al.(2021)Yan, Gao, Li, Zhang, Li, Huang, and
  Cui]{yan2021sparse}
Xu Yan, Jiantao Gao, Jie Li, Ruimao Zhang, Zhen Li, Rui Huang, and Shuguang
  Cui.
\newblock Sparse single sweep lidar point cloud segmentation via learning
  contextual shape priors from scene completion.
\newblock In \emph{Proceedings of the AAAI Conference on Artificial
  Intelligence}, pages 3101--3109, 2021.

\bibitem[Yang et~al.(2018)Yang, Wang, Stuckler, and Cremers]{yang2018deep}
Nan Yang, Rui Wang, Jorg Stuckler, and Daniel Cremers.
\newblock Deep virtual stereo odometry: Leveraging deep depth prediction for
  monocular direct sparse odometry.
\newblock In \emph{ECCV}, 2018.

\bibitem[Yu et~al.(2021)Yu, Ye, Tancik, and Kanazawa]{yu2021pixelnerf}
Alex Yu, Vickie Ye, Matthew Tancik, and Angjoo Kanazawa.
\newblock pixelnerf: Neural radiance fields from one or few images.
\newblock In \emph{CVPR}, pages 4578--4587, 2021.

\bibitem[Zamorski et~al.(2020)Zamorski, Zi{\k{e}}ba, Klukowski, Nowak, Kurach,
  Stokowiec, and Trzci{\'n}ski]{zamorski2020adversarial}
Maciej Zamorski, Maciej Zi{\k{e}}ba, Piotr Klukowski, Rafa{\l} Nowak, Karol
  Kurach, Wojciech Stokowiec, and Tomasz Trzci{\'n}ski.
\newblock Adversarial autoencoders for compact representations of 3d point
  clouds.
\newblock \emph{Computer Vision and Image Understanding}, 2020.

\bibitem[Zhan et~al.(2018)Zhan, Garg, Weerasekera, Li, Agarwal, and
  Reid]{zhan2018unsupervised}
Huangying Zhan, Ravi Garg, Chamara~Saroj Weerasekera, Kejie Li, Harsh Agarwal,
  and Ian Reid.
\newblock Unsupervised learning of monocular depth estimation and visual
  odometry with deep feature reconstruction.
\newblock In \emph{CVPR}, 2018.

\bibitem[Zhang et~al.(2019{\natexlab{a}})Zhang, Liu, Liu, and
  Huang]{zhang2019large}
Cheng Zhang, Zhi Liu, Guangwen Liu, and Dandan Huang.
\newblock Large-scale 3d semantic mapping using monocular vision.
\newblock In \emph{2019 IEEE 4th International Conference on Image, Vision and
  Computing (ICIVC)}, 2019{\natexlab{a}}.

\bibitem[Zhang et~al.(2018)Zhang, Zhao, Yao, Chen, Zhang, and
  Liao]{zhang2018efficient}
Jiahui Zhang, Hao Zhao, Anbang Yao, Yurong Chen, Li Zhang, and Hongen Liao.
\newblock Efficient semantic scene completion network with spatial group
  convolution.
\newblock In \emph{Proceedings of the European Conference on Computer Vision
  (ECCV)}, pages 733--749, 2018.

\bibitem[Zhang et~al.(2019{\natexlab{b}})Zhang, Liu, Lei, Lu, and
  Yang]{zhang2019cascaded}
Pingping Zhang, Wei Liu, Yinjie Lei, Huchuan Lu, and Xiaoyun Yang.
\newblock Cascaded context pyramid for full-resolution 3d semantic scene
  completion.
\newblock In \emph{Proceedings of the IEEE/CVF International Conference on
  Computer Vision}, pages 7801--7810, 2019{\natexlab{b}}.

\bibitem[Zhi et~al.(2021)Zhi, Laidlow, Leutenegger, and
  Davison]{Zhi2021semanticNERF}
Shuaifeng Zhi, Tristan Laidlow, Stefan Leutenegger, and Andrew~J. Davison.
\newblock In-place scene labelling and understanding with implicit scene
  representation.
\newblock In \emph{ICCV}, 2021.

\bibitem[Zhou et~al.(2021)Zhou, Greenwood, and Taylor]{zhou2021diffnet}
Hang Zhou, David Greenwood, and Sarah Taylor.
\newblock Self-supervised monocular depth estimation with internal feature
  fusion.
\newblock In \emph{BMVC}, 2021.

\bibitem[Zhou et~al.(2017)Zhou, Brown, Snavely, and Lowe]{zhou2017unsupervised}
Tinghui Zhou, Matthew Brown, Noah Snavely, and David~G Lowe.
\newblock Unsupervised learning of depth and ego-motion from video.
\newblock In \emph{CVPR}, 2017.

\end{thebibliography}
